\pdfoutput=1

\documentclass[11pt]{article}

\usepackage[final]{acl}
\usepackage{makecell}
\usepackage{graphicx}
\usepackage{array}
\usepackage{multirow}
\usepackage{multicol}
\usepackage{adjustbox}
\usepackage{booktabs}
\usepackage{siunitx}
\usepackage{placeins}
\usepackage{arydshln}  
\usepackage[skins,most]{tcolorbox} 
\tcbuselibrary{breakable} 
\newtcolorbox[auto counter, number within=section, list type=subsubsection, list inside=toc]{sectionbox}[2][]{
colback=white!98!gray, colframe=black, 
colbacktitle=white!90!gray, coltitle=black, 
fonttitle=\bfseries,
title={#2}, 
list entry={Comment \thetcbcounter\quad}
}
\usepackage{listings}
\lstdefinelanguage{json}{
    basicstyle=\ttfamily\small,
    breaklines=true,
    showstringspaces=false,
    frame=single,
    backgroundcolor=\color{gray!5},
    morestring=[b]",
    stringstyle=\color{blue},
    literate=
     *{0}{{{\color{black}0}}}{1}
      {1}{{{\color{black}1}}}{1}
      {2}{{{\color{black}2}}}{1}
      {3}{{{\color{black}3}}}{1}
      {4}{{{\color{black}4}}}{1}
      {5}{{{\color{black}5}}}{1}
      {6}{{{\color{black}6}}}{1}
      {7}{{{\color{black}7}}}{1}
      {8}{{{\color{black}8}}}{1}
      {9}{{{\color{black}9}}}{1},
}

\usepackage{booktabs}
\usepackage{subcaption}
\usepackage{times}
\usepackage{latexsym}
\usepackage{amsmath}
\usepackage{algorithm}
\usepackage{algpseudocode}
\usepackage{color}
\usepackage{listings}
\definecolor{customTeal}{RGB}{0, 128, 128}
\definecolor{emphasisColor}{RGB}{255, 0, 0} 
\usepackage{tabularx}
\usepackage{ragged2e}
\newcolumntype{Y}{>{\RaggedRight\arraybackslash}X} 
\usepackage[skins,most]{tcolorbox}
\usepackage{colortbl}
\usepackage{multirow} 
\usepackage{booktabs}
\usepackage{wrapfig}
\usepackage{graphicx}
\usepackage{tikz}
\FloatBarrier

\usepackage{pifont}  
\definecolor{customgreen}{HTML}{16C47F}  
\definecolor{customred}{HTML}{C62300}   

\newcommand{\cmark}{\textcolor{customgreen}{\ding{51}}}  
\newcommand{\xmark}{\textcolor{customred}{\ding{55}}}   

\usepackage{float}
\usepackage{hyperref}

\usepackage{caption}

\usepackage[T1]{fontenc}

\usepackage[utf8]{inputenc}

\usepackage{microtype}
\usepackage{afterpage}
\usepackage{inconsolata}
\usepackage{enumitem} 
\usepackage{lipsum}
\usepackage{graphicx}
\usepackage{pifont}
\definecolor{mycyan}{HTML}{008CBA}     
\definecolor{myyellow}{HTML}{B8860B}   
\definecolor{mylime}{HTML}{228B22}     

\NewDocumentCommand{\qingyun}
{ mO{} }{\textcolor{orange}{\textsuperscript{\textit{Qingyun}}\textsf{\textbf{\small[#1]}}}}

\NewDocumentCommand{\yi}
{ mO{} }{\textcolor{blue}{\textsuperscript{\textit{May}}\textsf{\textbf{\small[#1]}}}}

\NewDocumentCommand{\Mark}
{ mO{} }{\textcolor{green}{\textsuperscript{\textit{Minghao}}\textsf{\textbf{\small[#1]}}}}

\NewDocumentCommand{\zhitao}
{ mO{} }{\textcolor{gray}{\textsuperscript{\textit{Zhitao}}\textsf{\textbf{\small[#1]}}}}

\usepackage{titlesec}

\titlespacing{\subsection}{0pt}{*0}{*0}
\titlespacing{\subsubsection}{0pt}{*0}{*0}
\title{\raisebox{-0.3\height}{\includegraphics[height=3em]{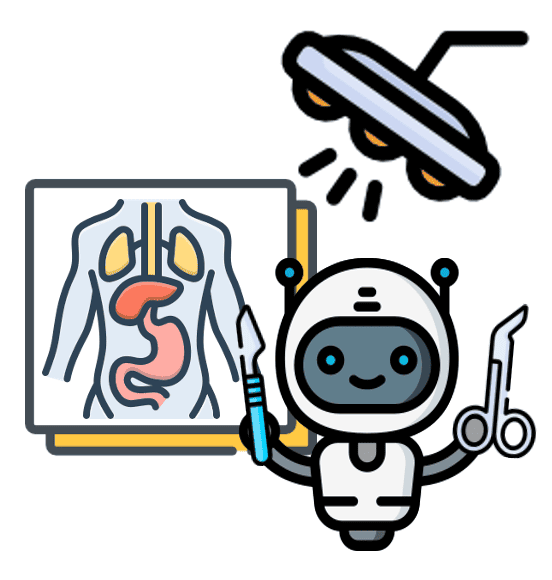}} MedEBench: Diagnosing Reliability in Text-Guided Medical Image Editing
}

\DeclareSymbolFont{extraup}{U}{zavm}{m}{n}
\DeclareMathSymbol{\vardiamond}{\mathalpha}{extraup}{87}
\author{\bf ~~~~Minghao Liu$^{\heartsuit}$ ~~~~~Zhitao He$^{\heartsuit}$ ~~~~~Zhiyuan Fan$^{\heartsuit}$ ~~~~~~Qingyun Wang$^{\diamondsuit}$ ~~~~~~Yi R. (May) Fung$^{\heartsuit}$\thanks{Corresponding author.}\\
$^{\heartsuit}$Hong Kong University of Science and Technology
~~~~~~~$^{\diamondsuit}$William \& Mary
\\
\texttt{mliuby@connect.ust.hk, yrfung@ust.hk}
}
\begin{document}
\widowpenalty=10000
\clubpenalty=10000

\maketitle

\vspace{-1em}

\begin{abstract}
Text-guided image editing has seen significant progress in natural image domains, but its application in medical imaging remains limited and lacks standardized evaluation frameworks. Such editing could revolutionize clinical practices by enabling personalized surgical planning, enhancing medical education, and improving patient communication. To bridge this gap, we introduce \textbf{MedEBench}\footnote{\url{https://mliuby.github.io/MedEBench_Website}}, a robust benchmark designed to diagnose reliability in text-guided medical image editing. MedEBench consists of 1,182 clinically curated image-prompt pairs covering 70 distinct editing tasks and 13 anatomical regions. It contributes in three key areas: (1) a clinically grounded evaluation framework that measures Editing Accuracy, Context Preservation, and Visual Quality, complemented by detailed descriptions of intended edits and corresponding Region-of-Interest (ROI) masks; (2) a comprehensive comparison of seven state-of-the-art models, revealing consistent patterns of failure; and (3) a diagnostic error analysis technique that leverages attention alignment, using Intersection-over-Union (IoU) between model attention maps and ROI masks to identify mislocalization issues, where models erroneously focus on incorrect anatomical regions. MedEBench sets the stage for developing more reliable and clinically effective text-guided medical image editing tools.
\end{abstract}
\section{Introduction}
Recent advances in diffusion models and vision-language pretraining have significantly advanced text-guided image editing~\cite{brooks2023instructpix2pixlearningfollowimage, kawar2023imagictextbasedrealimage, geng2023instructdiffusiongeneralistmodelinginterface, wasserman2025paintinpaintlearningadd, zhang2025incontexteditenablinginstructional, ge2025seedxmultimodalmodelsunified,google2024gemini2flash}. These methods enable diverse applications such as object removal, inpainting, and style transfer, often producing compelling visual results~\cite{yildirim2023instinpaint, wang2022imageneditor, yang2024freestyle}. However, most prior work has focused on aesthetic or creative tasks, with limited exploration in domains that require high semantic precision, particularly in medicine.
\begin{figure}[t]
\centering
\includegraphics[width=0.5\textwidth]{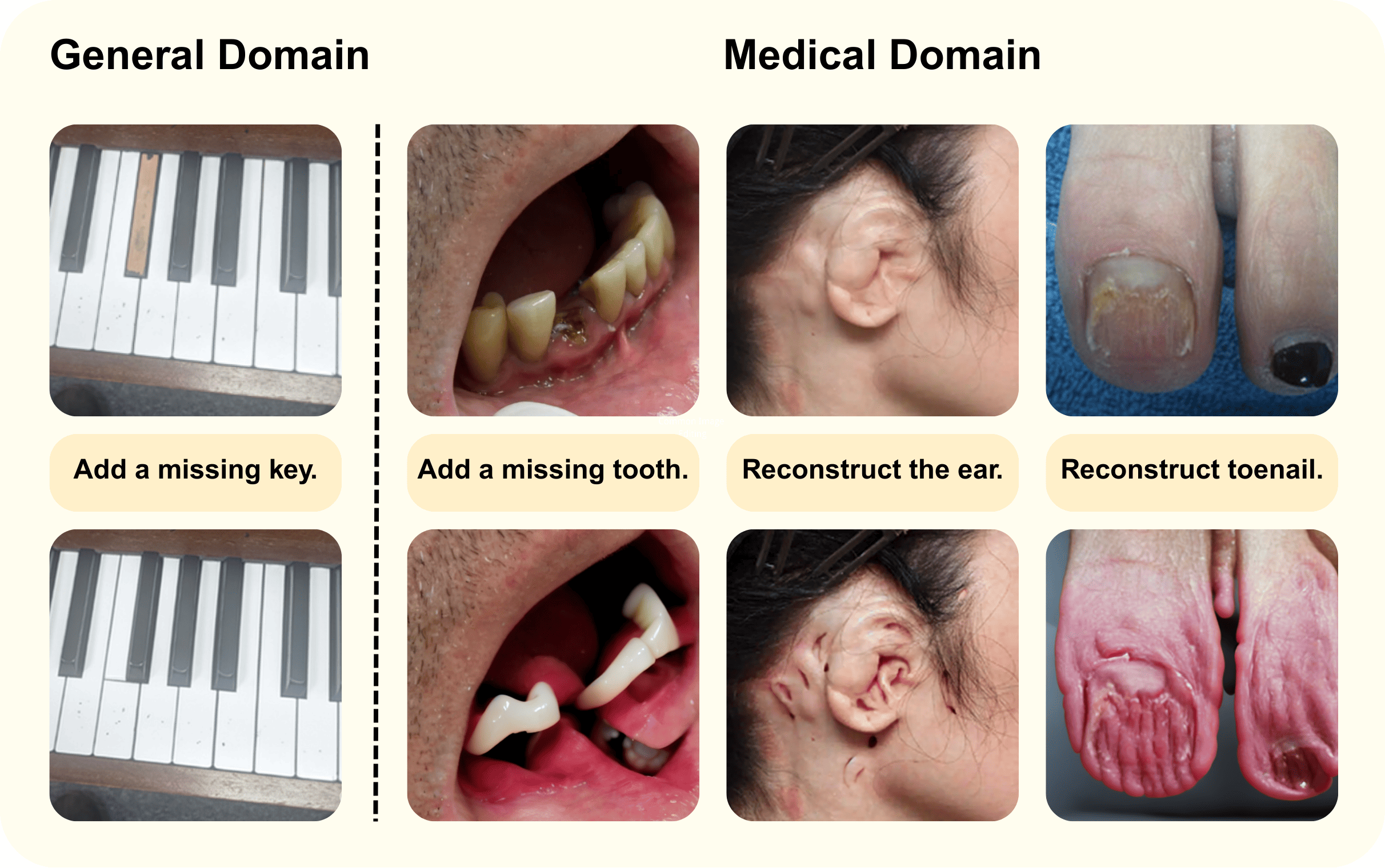} 
\caption{
State-of-the-Art model performs well on \textbf{common images} (e.g., “add a missing key”) but surprisingly struggles with \textbf{medical images} (e.g., “add a missing tooth”).
}
\label{fig:Figure1}
\end{figure}

\begin{figure*}[t]
\centering
\includegraphics[width=\textwidth]{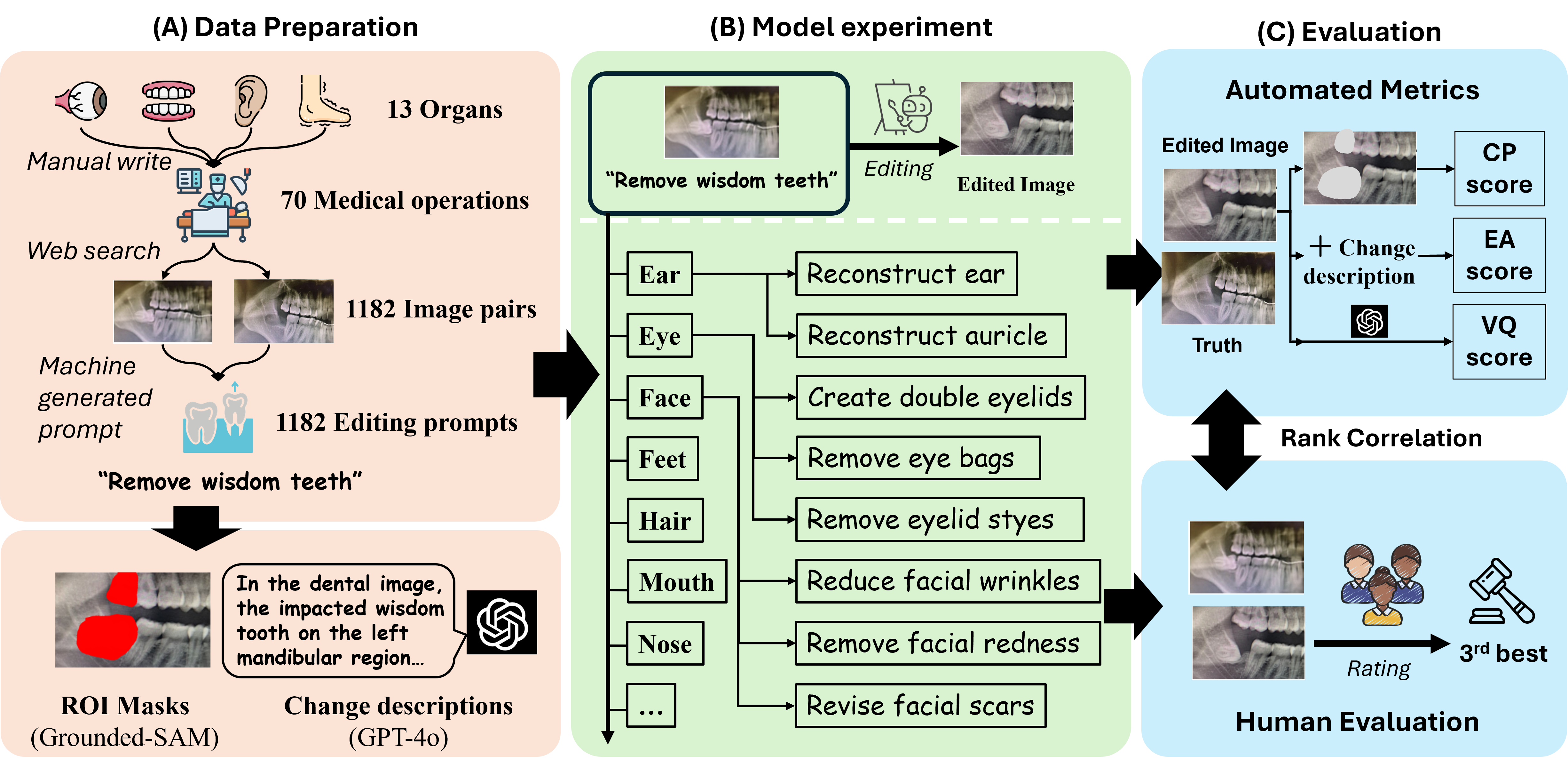}
\caption{Overview of \textbf{MedEBench}, a text-guided benchmark for medical image editing. (A) Data preparation includes collecting image triplets, generating ROI masks, and describing intended changes. (B) Models generate edited images from prompts and previous images. (C) SSIM structural similarity index measures contextual preservation; editing accuracy and visual quality are assessed by GPT-4o, guided by the change description.}
\label{fig:MedEBench}
\end{figure*}

While comprehensive benchmarks exist for evaluating editing models on natural images~\cite{lin2015microsoftcococommonobjects, huang2023t2icompbenchcomprehensivebenchmarkopenworld, xia2021tedigantextguideddiverseface}, their adaptation to medical images remains challenging. Here, even minor alterations can carry significant clinical meaning, demanding high editing precision, semantic fidelity, and anatomical correctness. Text-guided medical image editing holds substantial clinical potential, as it could highlight lesions in CT scans~\cite{guo2023generatect}, simulate surgical outcomes~\cite{huang2025skedittumor}, or generate personalized teaching materials~\cite{lee2024monet}. Such applications promise direct benefits to diagnosis, treatment optimization~\cite{ma2025textdiffseg}, and training~\cite{zhang2024aisd}. Despite the versatility of current models, they often fail at clinically meaningful transformations that are intuitive to general physicians or even non-experts. For example, InstructPix2Pix~\cite{brooks2023instructpix2pixlearningfollowimage} can successfully handle prompts such as ``Add a missing key on a piano keyboard'' but fails on similar medical prompts such as ``Add a missing tooth in this dental image'' (see Fig.~\ref{fig:Figure1}). Notably, this failure cannot be resolved by adjusting the text or image guidance scales~\cite{brooks2023instructpix2pix} (see Fig.~\ref{fig:labeled_axes}), where the \emph{text guidance scale} controls the classifier-free guidance weight for the text prompt, and the \emph{image guidance scale} controls the conditioning strength on the input image during denoising. This underscores the difficulty of transferring such models to domains that require fine-grained anatomical understanding and specialized domain knowledge~\cite{kazerouni2022diffusion, ayana2024multistage}.

Medical image editing faces two fundamental challenges. First, real image pairs that reflect clinically grounded transformations (e.g., before and after treatment) are scarce, and synthetic data often lacks realism. Second, evaluation remains problematic. Metrics such as Fréchet Inception Distance (FID)~\cite{heusel2018ganstrainedtimescaleupdate} and CLIP Score~\cite{hessel2022clipscorereferencefreeevaluationmetric} are not aligned with clinical correctness or anatomical plausibility, failing to capture if the edits are grounded in medical knowledge.  
To address these gaps, we propose \textbf{MedEBench}, a benchmark for text-guided medical image editing. MedEBench contains 1,182 real clinical image pairs covering pre- and postoperative states across 13 anatomical regions (e.g., teeth, eyes). Each edit case is defined by a natural language prompt, region-of-interest (ROI) masks, and detailed change descriptions to enable localized, fine-grained evaluation.
For clinically meaningful assessment, we introduce tailored evaluation metrics. \textit{Contextual Preservation} (CP) is measured by masked SSIM~\cite{wang2004ssim} to ensure unaffected regions remain intact. \textit{Editing Accuracy} (EA) and \textit{Visual Quality} (VQ) are assessed via Multimodal Large Language Models (MLLMs) using clinically detailed change descriptions for accurate, interpretable evaluation.
We benchmark seven state-of-the-art models across diverse learning paradigms and analyze failures through attention-grounding, revealing gaps in medical concept understanding and spatial localization. Our contributions are threefold:
\begin{itemize}[noitemsep,topsep=0pt,leftmargin=*]
    \item We introduce \textbf{MedEBench}, the first benchmark for text-guided medical image editing, featuring 1,182 real clinical image pairs from 13 anatomical regions and 70 editing tasks, each with prompts, ROI masks, and change descriptions.
    \item We propose clinically grounded evaluation metrics that capture structural preservation, edit accuracy, and visual quality, integrating both traditional approaches and MLLM-based reasoning.
    \item We benchmark seven models with text-instructed image editing ability on different learning paradigms and provide failure case analysis to diagnose limitations in medical concept editing by attention grounding. 
\end{itemize}
\begin{table*}[t]
\small
\centering
\begin{tabularx}{\textwidth}{l|c|c|c|>{\centering\arraybackslash}X}
\toprule
\textbf{Benchmark} & \textbf{Size} & \textbf{Domain} & \textbf{Synthetic} & \textbf{Truth} \\
\midrule
EditVal~\cite{basu2023editvalbenchmarkingdiffusionbased} & 648 pairs, 13 edit types & General & \xmark & \xmark \\
I2EBench~\cite{ma2024i2ebenchcomprehensivebenchmarkinstructionbased} & 2000+ pairs, 16 edit types & General & \xmark & mix \\
EditBench~\cite{wang2023imageneditoreditbenchadvancing} & 240 pairs & General & mix & \xmark \\
PIE-Bench~\cite{ju2023directinversionboostingdiffusionbased} & 700 pairs, 10 edit types & General & mix & \xmark \\
\midrule
\textbf{MedEBench (Ours)} & 1182 pairs, 13 organs, 70 types & \textbf{Medical} & \xmark & \cmark \\
\bottomrule
\end{tabularx}
\caption{Comparison of text-guided image editing benchmark datasets.}
\label{tab:image_editing_benchmarks}
\end{table*}

\section{MedEBench}

Fig.~\ref{fig:MedEBench} illustrates the main components of MedEBench. Section~\ref{sec:data-preparation} describes our dataset, which consists of 1,182 samples across 13 categories. Each sample includes an input image, a reference image, an editing prompt, an ROI mask, and a change description. Section~\ref{subsec:Automated_Evaluation} introduces automated evaluation metrics, while Section~\ref{subsec:human_evaluation} presents a human study validating the alignment between automated and human assessments. Tab.~\ref{tab:image_editing_benchmarks} compares MedEBench with existing benchmarks. As shown in Fig.~\ref{fig:Dataset_property}, low prompt-image CLIP similarity and varied ROI sizes highlight the challenges of text-guided medical image editing.

\subsection{Data Preparation}
\label{sec:data-preparation}
A major challenge in medical image editing is the scarcity of image pairs that reflect real clinical interventions. Unlike natural image datasets, synthesizing realistic transformations for medical images is difficult due to anatomical complexity and clinical constraints. 

\subsubsection{Image Pair Collection} 
\begin{figure}[h!]
\centering
\includegraphics[width=0.45\textwidth]{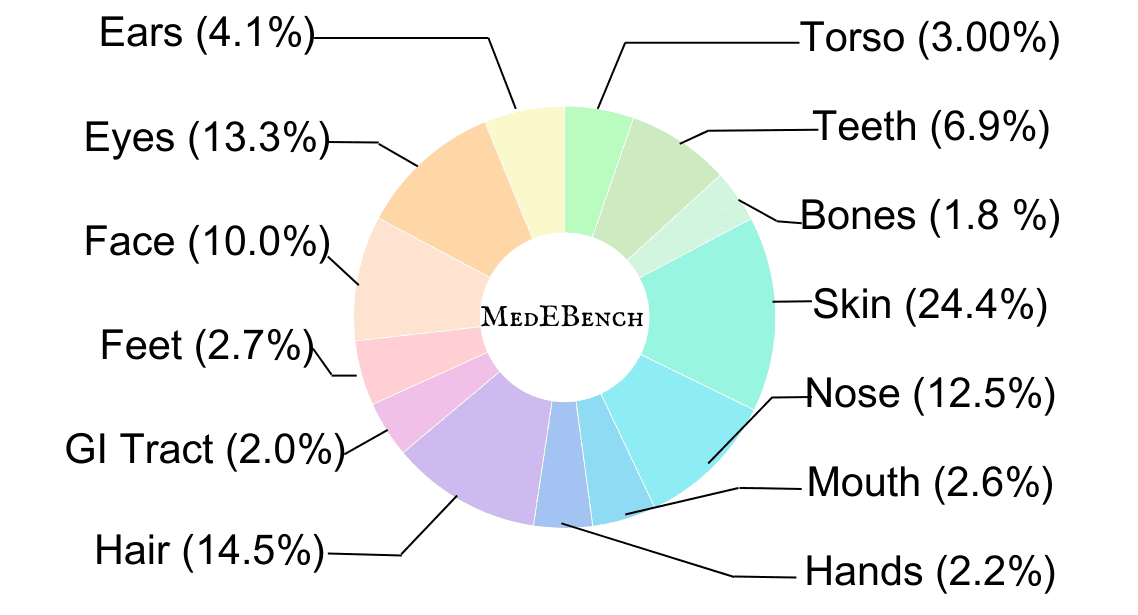}

\vspace{-1em}
\caption{Organ distribution}
\label{fig:Organ_counts}
\end{figure}

\begin{figure}[h!]
\centering
\includegraphics[width=0.49\textwidth]{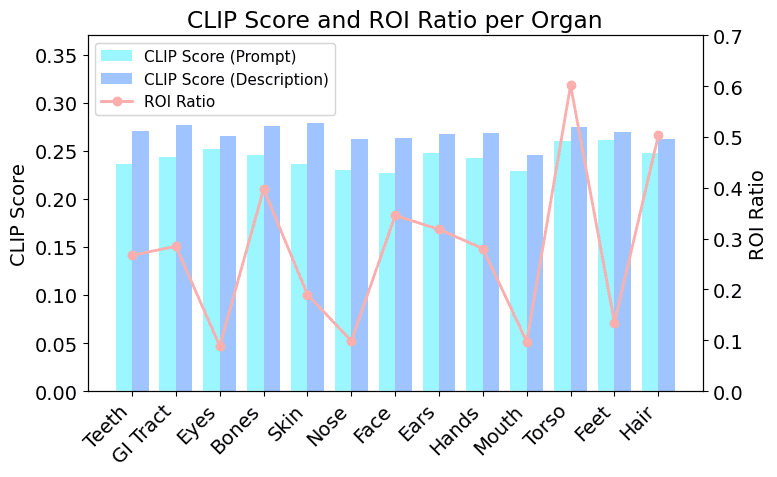}

\vspace{-1em}
\caption{Properties of the MedEBench dataset: CLIP score between the instruction and the preceding image, and the ROI mask ratio relative to the full image.}

\label{fig:Dataset_property}
\end{figure}

To mitigate data scarcity, we curate a set of ``before-and-after'' medical image pairs simulating realistic clinical transformations across various anatomical regions. We define a list of target organs $\mathcal{O}$ (e.g., \textit{Teeth}) and use ChatGPT~\cite{openai2024gpt4o} to generate corresponding medical procedures $\mathcal{A}_o$ (e.g., \textit{Remove wisdom teeth}), as shown in Tab.~\ref{tab:organ_task_sample_count} and Fig.~\ref{fig:Organ_counts}. Guided by these organ-procedure pairs, two expert annotators collected image pairs $(I_{\text{prev}}, I_{\text{after}})$ via keyword-based web search. As no existing dataset met our quality and alignment needs, they screened large volumes of candidates for anatomical accuracy, procedural relevance, and visual consistency. All pairs underwent two-stage review, were standardized to 512-pixel width, aligned via affine transformations, and anonymized for privacy.

\subsubsection{Editing Prompt Generation}
For each verified image pair $(I_{\text{prev}}, I_{\text{after}})$, we manually create a natural language prompt $p_{\text{orig}}$ describing the visual transformation, concisely reflecting the underlying medical operation. To enhance linguistic diversity and mitigate overfitting to fixed phrasings, we use ChatGPT~\cite{openai2024gpt4o} to generate multiple paraphrased variants. One variant is randomly selected as $p_{\text{reph}}$ for use in evaluation.

\subsubsection{Region-of-Interest Mask Annotation}
For each image $I_{\text{prev}}$, we generate a region-of-interest (ROI) mask $M$ to localize the area targeted by the editing prompt. Candidate masks are produced by prompting Grounded-SAM~\cite{ren2024groundedsamassemblingopenworld} with the instruction, yielding three proposals. A human annotator selects the most accurate mask or manually refines it to ensure anatomical precision. ROI masks serve two key purposes: (1) \textbf{Contextual Preservation}, where SSIM~\cite{wang2004ssim} is computed outside the masked region to verify that unedited areas remain unchanged; and (2) \textbf{Attention Analysis}, which evaluates whether model attention aligns with the relevant anatomy.

\subsubsection{Description of Change Generation}
\label{subsubsec:change_description}
Clinically meaningful evaluation requires more than visual fidelity or prompt alignment—it demands a precise understanding of anatomical changes. To this end, we generate structured \textit{descriptions of change} for each editing task, specifying the target anatomy and expected post-edit outcome. These task-specific descriptions replace generic prompts and guide Medical Large Language Model (MLLM)-based evaluations, enhancing interpretability and reliability. We employ GPT-4o~\cite{openai2024gpt4o} for its strong visual reasoning. For each sample, GPT-4o takes the input image pair and editing instruction, and generates a brief summary of the intended anatomical change. These descriptions support both automated and qualitative evaluation. See Appendix~\ref{appendix:gpt4o_prompt_description} for the full prompt.

\subsection{Automated Evaluation}
\label{subsec:Automated_Evaluation}
We develop an automated protocol to evaluate model performance across \textit{Contextual Preservation}, \textit{Editing Accuracy}, and \textit{Visual Quality}, combining traditional image metrics with GPT-4o-based judgment. A more detailed explanation and demonstration are provided in the Appendix.

\vspace{-0.5em}
\paragraph{Contextual Preservation} To evaluate whether the model preserves image regions unrelated to the intended edit, we compute SSIM\cite{wang2004ssim} between the previous image \(I_{\text{prev}}\) and the edited image \(I_{\text{edit}}\), excluding the region-of-interest (ROI) mask \(\mathcal{R}\). The contextual SSIM is defined as:
\(
\text{SSIM}_{\text{context}} = \text{SSIM}\left(I_{\text{prev}}|_{\overline{\mathcal{R}}}, I_{\text{edit}}|_{\overline{\mathcal{R}}}\right)
\). 
This metric captures how well the model maintains anatomical consistency outside the edited region.

\vspace{-0.5em}
\paragraph{Editing Accuracy and Visual Quality} We use GPT-4o~\cite{openai2024gpt4o}, a multimodal large language model with visual reasoning capabilities, to evaluate \textit{Editing Accuracy} and \textit{Visual Quality}. For each sample, GPT-4o is provided with the description of change, previous image, edited image, and ground truth image, and follows a structured two-step protocol:

\vspace{-0.5em}
\paragraph{Step 1: Visual Difference Description.}
GPT-4o first compares the previous and edited images to describe all visible changes, identifying what has been added, removed, or modified, along with the anatomical regions affected. 

\vspace{-0.5em}
\paragraph{Step 2: Scoring.}
Guided by the reference \textit{description of change} generated for the task, GPT-4o evaluates the following aspects: 
\textbf{Editing Accuracy (0--10)} measures how well the actual changes in the edited image match the expected transformation described in the reference, reflecting completeness and correctness with deductions for irrelevant or missing edits; 
\textbf{Visual Quality (0--10)} assesses the realism, clarity, and overall visual fidelity of the edited image. 
Each score is accompanied by a concise rationale to enhance transparency and evaluation reliability.

\section{Experiment and Results}
\subsection{Baseline Models}
We evaluate seven state-of-the-art models with text-guided image editing capabilities. TIE models: 1) \textbf{InstructPix2Pix}~\cite{brooks2023instructpix2pixlearningfollowimage}: An early diffusion-based model fine-tuned on synthetic instruction-image pairs for prompt-based editing with strong spatial alignment.  2) \textbf{Imagic}~\cite{kawar2023imagictextbasedrealimage}: Optimizes latent codes to enable realistic edits of real images, preserving identity and structure without requiring paired data.  3) \textbf{InstructDiffusion}~\cite{geng2023instructdiffusiongeneralistmodelinginterface}: A generalist model for instruction-following across diverse vision tasks, supporting flexible zero-shot editing. 4) \textbf{Paint-by-Inpaint}~\cite{wasserman2025paintinpaintlearningadd}: Proposes an object addition paradigm via region removal and inpainting-based completion, enabling mask-free object insertion.  
5) \textbf{ICEdit}~\cite{zhang2025incontexteditenablinginstructional}: Utilizes diffusion transformers (DiT) with in-context learning and adapter tuning for few-shot instructional editing.  Universal MLLMs:
6) \textbf{SEED-X}~\cite{ge2025seedxmultimodalmodelsunified}: A unified multimodal model supporting both image understanding and generation for general-purpose editing. 7) \textbf{Gemini 2.0 Flash}~\cite{google2025gemini}: A commercial-grade multimodal system integrating fast image generation, conversational interaction, and robust editing capabilities. To ensure fair comparison, we perform hyperparameter sweeps for each model around default configurations. Detailed settings and prompts are provided in the Appendix.

\begin{table*}[htbp]

\centering
\begin{adjustbox}{max width=\textwidth}
\begin{tabular}{|l*{7}{|c@{\hskip 4pt}c@{\hskip 4pt}c}|}
\toprule
\multirow{2}{*}{\textbf{Model}} &
\multicolumn{3}{c|}{\textbf{Teeth}} &
\multicolumn{3}{c|}{\textbf{Eyes}} &
\multicolumn{3}{c|}{\textbf{Spine}} &
\multicolumn{3}{c|}{\textbf{Skin}} &
\multicolumn{3}{c|}{\textbf{Nose}} &
\multicolumn{3}{c|}{\textbf{Face}} &
\multicolumn{3}{c|}{\textbf{GI Tract}} \\
\cmidrule(lr){2-4} \cmidrule(lr){5-7} \cmidrule(lr){8-10}
\cmidrule(lr){11-13} \cmidrule(lr){14-16} \cmidrule(lr){17-19} \cmidrule(lr){20-22}
& EA & CP & VQ & EA & CP & VQ & EA & CP & VQ & EA & CP & VQ & EA & CP & VQ & EA & CP & VQ & EA & CP & VQ \\
\midrule
imagic & \underline{0.49} & 0.64 & 0.57 & 0.50 & 0.49 & 0.62 & 0.18 & 0.66 & \underline{0.52} & 0.19 & 0.53 & 0.37 & 0.40 & 0.40 & 0.48 & 0.27 & 0.55 & 0.42 & 0.13 & 0.63 & 0.58 \\
instruct-pix2pix & 0.32 & \textbf{0.85} & 0.42 & 0.59 & \textbf{0.90} & \textbf{0.75} & 0.21 & \underline{0.79} & 0.36 & 0.51 & \textbf{0.85} & 0.65 & 0.66 & \textbf{0.86} & \underline{0.72} & 0.70 & \underline{0.85} & \textbf{0.79} & 0.19 & \textbf{0.87} & 0.67 \\
instruct-diffusion & 0.47 & 0.67 & \underline{0.65} & \textbf{0.64} & 0.66 & \underline{0.72} & 0.06 & 0.72 & \underline{0.52} & 0.56 & 0.69 & 0.59 & 0.59 & 0.57 & 0.69 & 0.54 & 0.68 & 0.58 & 0.18 & 0.62 & 0.45 \\
paint-by-inpaint & 0.19 & 0.67 & 0.25 & 0.40 & 0.57 & 0.43 & 0.14 & 0.67 & 0.27 & 0.25 & 0.47 & 0.30 & 0.54 & 0.61 & 0.54 & 0.36 & 0.54 & 0.36 & 0.10 & 0.57 & 0.33 \\
icedit & 0.46 & 0.79 & 0.57 & \underline{0.60} & 0.72 & 0.69 & 0.16 & 0.67 & 0.50 & \underline{0.64} & 0.78 & \underline{0.72} & \underline{0.72} & \underline{0.77} & \textbf{0.75} & \underline{0.71} & 0.78 & \underline{0.76} & \underline{0.37} & 0.78 & \underline{0.70} \\
\cdashline{1-1}
seedx & 0.32 & 0.70 & 0.47 & 0.33 & \underline{0.79} & 0.55 & \underline{0.37} & 0.65 & 0.48 & 0.26 & 0.79 & 0.54 & 0.24 & \underline{0.77} & 0.57 & 0.33 & 0.76 & 0.58 & 0.22 & 0.61 & 0.49 \\
gemini\_2\_flash & \textbf{0.75} & \underline{0.84} & \textbf{0.81} & 0.55 & 0.78 & 0.63 & \textbf{0.38} & \textbf{0.81} & \textbf{0.76} & \textbf{0.77} & \underline{0.84} & \textbf{0.77} & \textbf{0.77} & 0.76 & \underline{0.72} & \textbf{0.72} & \textbf{0.86} & 0.70 & \textbf{0.66} & \underline{0.80} & \textbf{0.76} \\
\end{tabular}
\end{adjustbox}
\begin{adjustbox}{max width=\textwidth}
\begin{tabular}{
|c@{\hskip 4pt}c@{\hskip 4pt}c|
 c@{\hskip 4pt}c@{\hskip 4pt}c|
 c@{\hskip 4pt}c@{\hskip 4pt}c|
 c@{\hskip 4pt}c@{\hskip 4pt}c|
 c@{\hskip 4pt}c@{\hskip 4pt}c|
 c@{\hskip 4pt}c@{\hskip 4pt}c|
 >{\hskip 2pt}c<{\hskip 0pt} 
 c@{\hskip 4pt}c@{\hskip 4pt}c@{\hskip 4pt}c>{\hskip 8pt}c<{\hskip 0pt}| 
}
\toprule
\midrule
\multicolumn{3}{|c|}{\textbf{Ears}} &
\multicolumn{3}{c|}{\textbf{Hands}} &
\multicolumn{3}{c|}{\textbf{Mouth}} &
\multicolumn{3}{c|}{\textbf{Torso}} &
\multicolumn{3}{c|}{\textbf{Feet}} &
\multicolumn{3}{c|}{\textbf{Hair}} &
{} &
\multicolumn{4}{c|}{\textbf{Overall}} \\
\cmidrule(lr){1-3}
\cmidrule(lr){4-6}
\cmidrule(lr){7-9}
\cmidrule(lr){10-12}
\cmidrule(lr){13-15}
\cmidrule(r){16-18}
\cmidrule(lr){20-23}
EA & CP & VQ & EA & CP & VQ & EA & CP & VQ & EA & CP & VQ & EA & CP & VQ & EA & CP & VQ & {} & EA & CP & VQ & FID& {} \\
\midrule
0.40 & 0.58 & 0.50 & 0.41 & 0.51 & 0.45 & 0.40 & 0.43 & 0.39 & 0.33 & 0.67 & 0.51 & 0.26 & 0.55 & 0.38 & 0.64 & 0.71 & 0.53 & {} & 0.38 & 0.55 & 0.48 & 88.54& {} \\
0.36 & \textbf{0.93} & 0.66 & 0.36 & \underline{0.75} & 0.39 & 0.64 & \underline{0.77} & 0.64 & 0.18 & 0.73 & 0.35 & 0.17 & \textbf{0.90} & 0.48 & 0.42 & \underline{0.88} & 0.44 & {} & 0.50 & \textbf{0.86} & 0.62 & \underline{46.50} & {}\\
0.45 & 0.75 & 0.63 & 0.49 & 0.66 & 0.59 & 0.61 & 0.51 & 0.63 & 0.31 & 0.67 & 0.54 & 0.23 & 0.75 & 0.74 & \underline{0.68} & 0.70 & \underline{0.65} & {} & 0.54 & 0.67 & 0.63 & 65.56 & {}\\
0.25 & 0.66 & 0.33 & 0.40 & 0.67 & 0.40 & 0.41 & 0.59 & 0.43 & 0.35 & 0.66 & 0.30 & 0.16 & 0.58 & 0.20 & 0.60 & 0.71 & 0.51 & {} & 0.37 & 0.59 & 0.38 & 109.61 & {}\\
\underline{0.70} & \underline{0.83} & \underline{0.76} & \underline{0.50} & 0.70 & \underline{0.68} & \underline{0.71} & 0.71 & \underline{0.74} & \underline{0.62} & \textbf{0.90} & \textbf{0.74} & \underline{0.74} & 0.84 & \underline{0.76} & \textbf{0.71} & \textbf{0.90} & \textbf{0.79} & {} & \underline{0.60} & 0.79 & \textbf{0.72} & 46.76 & {}\\
0.41 & 0.81 & 0.51 & 0.43 & 0.74 & 0.54 & 0.27 & \textbf{0.80} & 0.67 & 0.35 & 0.72 & 0.50 & 0.07 & \underline{0.88} & 0.56 & 0.40 & 0.86 & 0.52 & {} & 0.31 & 0.78 & 0.54 & 51.04& {} \\
\textbf{0.76} & 0.72 & \textbf{0.82} & \textbf{0.82} & \textbf{0.81} & \textbf{0.79} & \textbf{0.73} & 0.72 & \textbf{0.79} & \textbf{0.62} & \underline{0.87} & \textbf{0.74} & \textbf{0.82} & 0.86 & \textbf{0.82} & 0.50 & 0.87 & 0.63 & {} & \textbf{0.68} & \underline{0.82} & \textbf{0.72} & \textbf{43.74} & {}\\

\bottomrule
\end{tabular}
\end{adjustbox}
\caption{
Editing performance across organs. \textbf{Bold} = best, \underline{underline} = second best. 
EA = GPT-4o Editing Accuracy$_\text{Desc}$, CP = Masked SSIM, VQ = GPT-4o Visual Quality$_\text{Desc}$. 
EA and VQ scaled from 0–10 to 0–1.
}
\label{tab:organ_benchmark}
\end{table*}
\begin{figure*}[h]
\centering
\includegraphics[width=\textwidth]{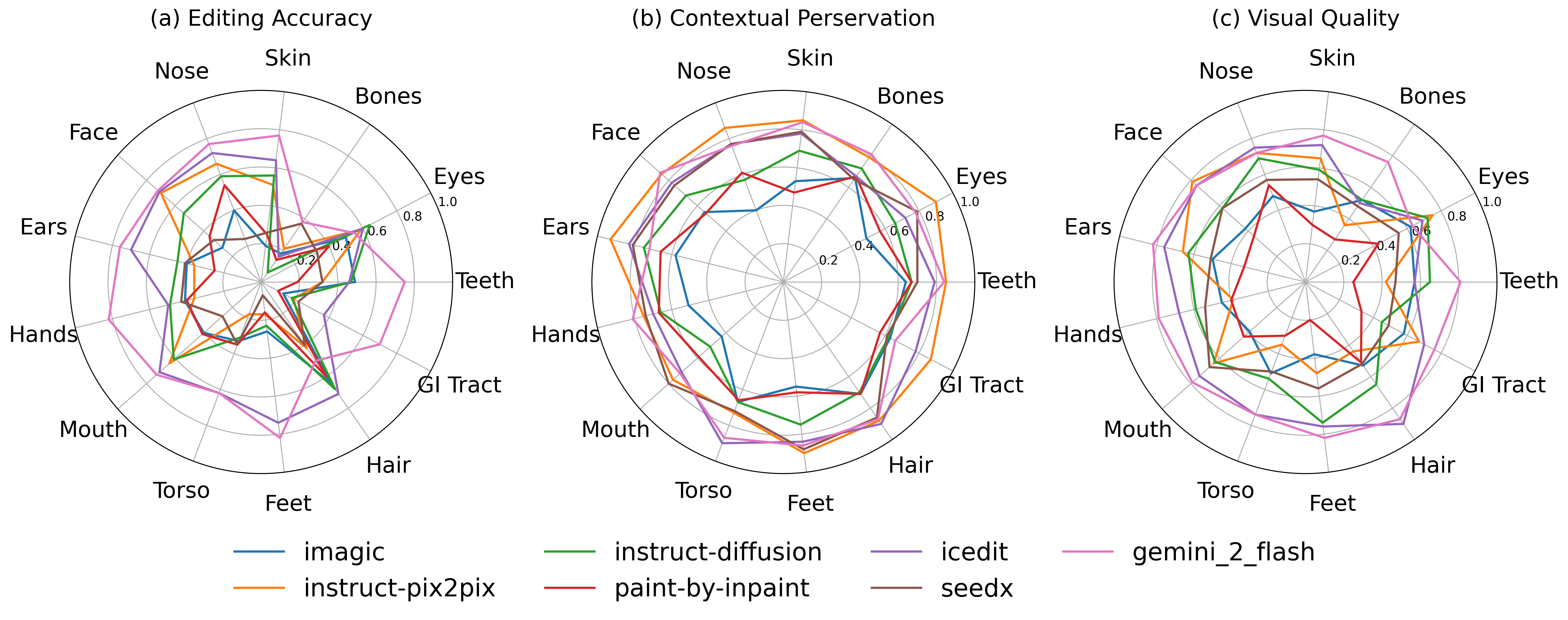}
\vspace{-0.2em}
\caption{Per-organ performance comparison of seven image editing models across three metrics.}
\label{fig:radar}
\end{figure*}

\subsection{Baseline Metrics}
\label{subsec:baseline_metrics}
We compare our proposed metrics (Section~\ref{subsec:Automated_Evaluation}) with commonly used automated metrics for image editing quality assessment, focusing on their correlation with human evaluation results (Section~\ref{subsec:human_evaluation}). For CLIP-based metrics: 1) \textbf{ISim} measures the similarity between the edited image and the ground truth in the CLIP embedding space; 2) \textbf{TAlign} evaluates the alignment between text instructions and edited images; 3) \textbf{DAlign} captures the directional consistency of edits with respect to text guidance. In the Pixel Similarity category, traditional image quality metrics compare edited images against ground truth: 4) \textbf{PSNR}~\cite{korhonen2012psnr} measures pixel-wise reconstruction accuracy; 5) \textbf{LPIPS}~\cite{zhang2018lpips} quantifies perceptual similarity using deep visual features; and 6) \textbf{SSIM}~\cite{wang2004ssim} assesses structural similarity. Additionally, the reward-based metric 7) \textbf{ImageReward}~\cite{xu2023imagereward} provides a learned perceptual score designed to correlate with human preferences. Finally, 8) \textbf{FID}~\cite{heusel2018ganstrainedtimescaleupdate} measures the distance between the distributions of real and generated images in a feature space. Since FID is computed at the distribution level, it is not included in our correlation comparison but is reported as a reference in Section~\ref{subsec:main_results}.

\subsection{Main Results}
\label{subsec:main_results}
We summarize model performance on MedEBench in Tab.~\ref{tab:organ_benchmark} and visualize key trends in Fig.~\ref{fig:radar}. Representative editing examples are shown in Tab.~\ref{tab:editing_table}. Gemini 2 Flash achieves the best overall performance, leading in editing accuracy (EA = 0.68), visual quality (VQ = 0.72, tied with ICEdit), and ranking second in context preservation (CP = 0.82). It also delivers the most realistic outputs, as reflected by its superior Fréchet Inception Distance (FID = 43.74)~\cite{heusel2018ganstrainedtimescaleupdate}. Among open-source models, ICEdit shows the most balanced performance (EA = 0.60, VQ = 0.72, CP = 0.79). InstructPix2Pix, while excelling in context preservation (CP = 0.86), suffers from lower editing accuracy (EA = 0.50), likely due to its reluctance to apply medically significant edits. A substantial gap persists between Gemini and open-source methods with over 10\% in editing accuracy, highlighting limitations of current  methods for high-precision medical editing.

\textbf{At the organ level}, editing difficulty varies significantly. Regions such as the \textit{spine} and \textit{bones} (e.g., CT scans) remain especially challenging, with the best EA only reaching 0.38. This difficulty is not primarily due to grayscale input but rather the inherent \textit{structural complexity} of bones. Unlike soft tissues, bones follow rigid, geometrically consistent patterns (e.g., aligned vertebrae, symmetric ribs), making even small editing errors visually salient and disruptive. For example, in a task like \textit{“correct spines”}, the model must straighten the spine without disturbing rib symmetry or vertebral alignment, demanding a much higher degree of structural precision than tasks such as \textit{“remove intestinal polyps”}. The largest performance gaps between Gemini and open-source models are seen in \textit{teeth} (34\%), \textit{hands} (64\%), and \textit{gastrointestinal tract} (43\%), further reflecting the challenges posed by complex anatomical structures and repetitive patterns. Conversely, superficial structures such as \textit{hair} and \textit{nose} are easier to edit, where ICEdit achieves strong performance (hair: EA = 0.71, VQ = 0.79).

\textbf{Model-specific strengths} are observed. Gemini 2.0 Flash excels at precision-demanding edits and complex internal organ modifications. For instance, in the “Remove or beautify rhinophyma” task, it is the only model that successfully smooths nodular irregularities, reduces erythema, and reshapes the nose, aided by Dynamic Prompt Restatement~\cite{google2025gemini} that reformulates user instructions into more detailed, context-aware prompts. Its targeted, incremental editing strategy further enhances realism, though Gemini sometimes repaints large regions (e.g., mouth and eyes), introducing subtle artifacts that risk compromising clinical reliability. ICEdit, powered by DiT-based in-context editing, handles complex anatomy effectively but struggles with concept removal (e.g., \textit{teeth} edits). SEED-X shows surprising strength in \textit{spine} edits (EA = 0.37), outperforming other open-source models, possibly due to better alignment with CT image modalities.

Three key observations arise: \textbf{(1) Internal organs} are significantly harder to edit than superficial structures, with EA scores averaging 2.3$\times$ lower; \textbf{(2) Gemini’s advantage} is most pronounced in anatomically complex regions with repetitive patterns (e.g., hands, gastrointestinal tract); and \textbf{(3) A trade-off exists} between visual quality and context preservation: models with higher VQ, such as Gemini, tend to exhibit slightly lower CP, highlighting the need for localized, precise edits rather than full-image redraws in medical applications.

\subsection{Learning Paradigms Comparison}
\label{subsec:Learning_Paradigm_Comparison_Results}
To explore different learning paradigms for medical image editing, we select six representative tasks and sample 30 images per task. We compare two representative approaches: \textbf{Fine-tuning.} We adopt InstructPix2Pix as a representative fine-tuning method. Its U-Net backbone is fine-tuned using triplets of input images, editing instructions, and ground-truth edited outputs. Training is performed for 50 steps per sample. \textbf{In-context learning.} We evaluate Gemini's in-context capability via few-shot prompting. During inference, prompts are constructed using several demonstration triplets (previous image, editing instruction, and ground-truth), followed by a test image and instruction.

As shown in Tab.~\ref{tab:Learning_Paradigms_result}, fine-tuning proves effective for InstructPix2Pix, particularly in reconstruction tasks. Editing accuracy improves consistently with more fine-tuning samples. However, tasks such as \textit{Remove Wisdom Teeth} and \textit{Remove Moles} show diminishing returns, with accuracy gains plateauing and often compromising contextual preservation.

In contrast, Gemini's in-context learning shows limited effectiveness. Increasing the number of demonstrations does not improve accuracy and often degrades performance. The model struggles to distinguish between test inputs and in-context examples, leading to confusion and reduced contextual consistency. These findings highlight the difficulty of applying in-context learning to fine-grained, pixel-level medical editing.

\begin{table}[htbp]
\small
\centering
\begin{tabularx}{\columnwidth}{>{\raggedright\arraybackslash}p{3cm}| S[table-format=2.2] S[table-format=2.2] S[table-format=2.2]}
\toprule
\textbf{Metric} & \textbf{EA} & \textbf{CP} & \textbf{VQ} \\
\midrule

\multirow{5}{*}{\rotatebox{90}{CLIP}} 
ISim               & 0.35 & 0.55 & 0.56 \\
\quad TAlign$_{\text{P}}$          & 0.16 & 0.26 & 0.28 \\
\quad TAlign$_{\text{D}}$         & 0.22 & 0.36 & 0.33 \\
\quad DAlign$_{\text{P}}$          & 0.15 & 0.25 & 0.24 \\
\quad DAlign$_{\text{D}}$      & 0.20 & 0.35 & 0.32 \\
\midrule

\multirow{6}{*}{\rotatebox{90}{Pixel/Reward}} 
PSNR               & 0.33 & 0.65 & 0.54 \\
\quad LPIPS              & 0.38 & 0.67 & 0.60 \\
\quad SSIM               & 0.20 & 0.66 & 0.43 \\
\quad \textbf{Masked SSIM}        & 0.21 & \textbf{ 0.82} & 0.45 \\
\quad ImageReward$_{\text{P}}$    & 0.20 & 0.28 & 0.41 \\
\quad ImageReward$_{\text{D}}$  & 0.16 & 0.27 & 0.36 \\
\midrule

\multirow{6}{*}{\rotatebox{90}{GPT-4o}} 
Accuracy$_{\text{P}}$       & 0.64 & 0.21 & 0.42 \\
\quad \textbf{Accuracy}$_{\text{D}}$      & \textbf{ 0.79} & 0.18 & 0.46 \\
\quad Context$_{\text{P}}$       & 0.41 & 0.56 & 0.69 \\
\quad Context$_{\text{D}}$       & 0.43 & 0.57 & 0.76 \\
\quad Quality$_{\text{P}}$       & 0.44 & 0.44 & 0.81 \\
\quad \textbf{Quality$_{\text{D}}$} & 0.46 & 0.46 & \textbf{ 0.82} \\
\bottomrule
\end{tabularx}
\caption{
Spearman Rank Correlation ($\rho$) Between Human Ratings (column) and Automated Metrics (row).
}
\label{tab:spearman_filled}
\end{table}
\subsection{Human Evaluation} \label{subsec:human_evaluation}
To validate our automated metrics, we conduct a human evaluation study based on relative ranking. Two expert annotators independently assess a subset of edited images across three dimensions: \textit{Editing Accuracy} (EA), \textit{Contextual Preservation} (CP), and \textit{Visual Quality} (VQ), consistent with the automated evaluation framework. The evaluation panel and annotator instructions appear in Fig.~\ref{fig:human_eval_panel}. For each organ $o \in \mathcal{O}$, we randomly sample 20 editing tasks, yielding a total of 260 edited samples. Outputs from multiple models are collected using identical inputs and prompts. Annotators rank the model outputs for each sample along all three dimensions. The inter-annotator Spearman rank correlation coefficient reaches 0.91, indicating excellent agreement and high annotation reliability. 

To assess the consistency between human and automated evaluations, we compute the Spearman correlation coefficient $\rho$~\cite{spearman1904proof} between human-assigned ranks $R_h(i)$ and automated ranks $R_a(i)$ for each sample $i$ across $m$ models:
\begin{equation}
\rho = 1 - \frac{6 \sum_i (R_h(i) - R_a(i))^2}{m(m^2 - 1)}
\end{equation}
We evaluate both baseline metrics (Section~\ref{subsec:baseline_metrics}) and our proposed ones (Section~\ref{subsec:Automated_Evaluation}). To further investigate the impact of prompt structure, we introduce alternative versions of each text-based metric by replacing the original editing prompts with structured \textit{descriptions of change} (Section~\ref{subsubsec:change_description}). As shown in Tab.~\ref{tab:spearman_filled}, our proposed metrics (Masked SSIM for CP, GPT-4o based EA, and VQ with detailed change descriptions) achieve the highest alignment with human assessments across all three evaluation dimensions. Notably, incorporating expected change guidance into GPT-4o scoring improves correlation with expert judgments by 23\%.

\section{Failure Analysis via Attention}
We select the InstructPix2Pix model as a case study for failure analysis, as it is a widely used text-guided image editing model known for strong context preservation but relatively low editing accuracy, as discussed in Section~\ref{subsec:main_results}. Notably, InstructPix2Pix demonstrates a tendency to preserve global image structures while failing to execute fine-grained edits, making it an ideal candidate for analyzing attention-related failures. Prior work~\cite{liu2024understandingcrossselfattentionstable} has shown that cross-attention mechanisms in diffusion models play a critical role in localizing prompt tokens to corresponding image regions, and in encoding semantic and categorical information. To further investigate this phenomenon, we examine whether the editing process attends to the correct anatomical regions by analyzing the model's cross-attention maps. For each editing prompt, we extract the cross-attention maps corresponding to the last token of the key visual concept \( t_c \) across all diffusion steps. These are averaged to obtain a single attention vector:
\(
\bar{\mathbf{a}}_{t_c} = \frac{1}{S} \sum_{s=1}^{S} \mathbf{a}^{(s)}_{t_c}
\). 
This vector is reshaped into a 2D map \( \bar{A}_{t_c} \), normalized to the range \([0, 1]\), and binarized using a scaled (We choose \(\alpha = 1.3\).) Otsu threshold~\cite{otsu1979threshold}:
\begin{equation}
A = \left( \frac{\bar{A}_{t_c} - \min}{\max - \min} > \alpha \cdot \tau_{\text{Otsu}} \right)  
\end{equation}
We then compare the thresholded attention map \( A \) with the annotated Region-of-Interest (ROI) mask \( M \) using the Intersection-over-Union: \(
\text{IoU} = |A \cap M|/|A \cup M|
\). A higher IoU score indicates stronger spatial alignment between the model’s attention and the intended anatomical target, reflecting better grounding of visual concepts during editing. The average IoU scores for each organ are listed in Tab.~\ref{tab:attention_iou_by_organ}.
\begin{table}[ht]
\small
\centering
\begin{tabular}{l c | l c | l c}
\toprule
\textbf{Organ} & \textbf{IoU} & \textbf{Organ} & \textbf{IoU} & \textbf{Organ} & \textbf{IoU}\\
\midrule

\small{Feet}  & 0.189 & \small{Teeth} & 0.268 & GI Tract & 0.4084 \\
\small{Skin} & 0.191 & \small{Face}  & 0.289 & Hands & 0.409 \\
\small{Nose} & 0.193 & \small{Bones} & 0.389 & Hair & 0.497\\
\small{Eyes} & 0.195  & \small{Ears} & 0.407  & Torso & 0.592\\
\small{Mouth} & 0.224 &  -   &      -  &   -    &   -     \\
\bottomrule
\end{tabular}
\caption{Average IoU between attention maps and ROI masks across anatomical regions in InstructPix2Pix.}

\label{tab:attention_iou_by_organ}
\end{table}

\vspace{-1em}
\paragraph{Analysis}  
We analyze failure cases characterized by \textbf{low Editing Accuracy (EA)} and \textbf{high Context Preservation (CP)}, indicating that edits were insufficient or absent, though the overall image remained intact. In this setting  IoU between attention maps and ground-truth regions serves as a diagnostic indicator to distinguish failure types:
\begin{figure}[h]
    \centering
    \begin{subfigure}[t]{0.45\textwidth}
        \centering
        \begin{tikzpicture}
            \node[anchor=south west, inner sep=0] (img) at (0,0)
                {\includegraphics[width=\textwidth]{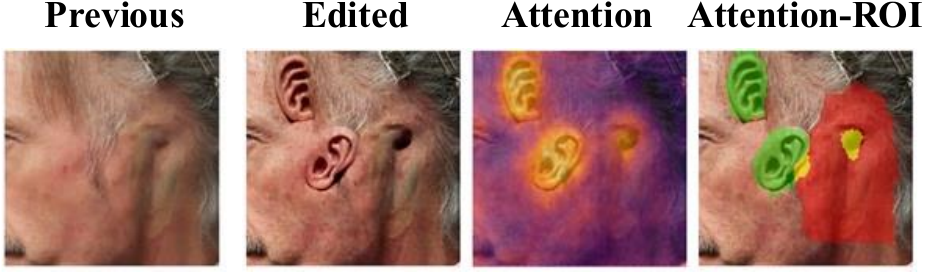}};
            \draw[black, ultra thick, dashed] (0.25\textwidth, 2.1) -- (0.25\textwidth, 0); 
        \end{tikzpicture}
        \caption*{\small
        (1)
        \textbf{Prompt:} Reconstruct \underline{ear}. 
        \textbf{IoU:} 0.04}
    \end{subfigure}
    \begin{subfigure}[t]{0.45\textwidth}
        \centering
        \begin{tikzpicture}
            \node[anchor=south west, inner sep=0] (img) at (0,0)
                {\includegraphics[width=\textwidth]{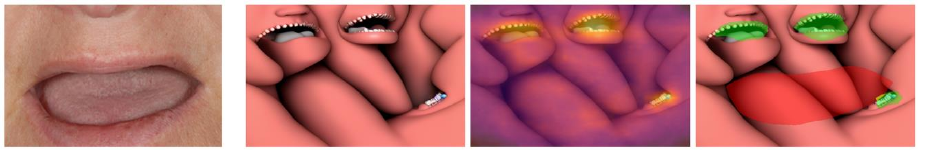}};
            \draw[black, ultra thick, dashed] (0.25\textwidth, 1.08) -- (0.25\textwidth, 0); 
        \end{tikzpicture}
        \caption*{\small 
        (2)
        \textbf{Prompt:} Implant \underline{teeth}. \textbf{IoU:} 0.003}
    \end{subfigure}

        \begin{subfigure}[t]{0.45\textwidth}
        \centering
        \begin{tikzpicture}
            \node[anchor=south west, inner sep=0] (img) at (0,0)
                {\includegraphics[width=\textwidth]{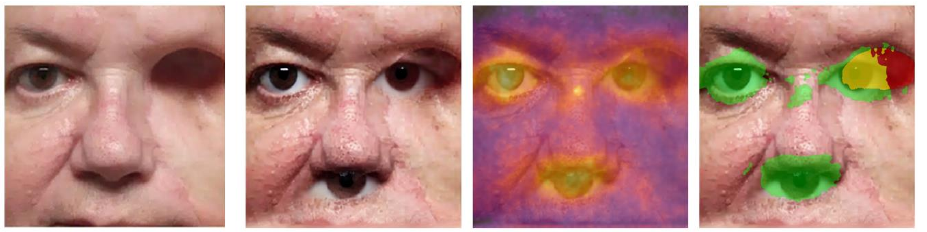}};
            \draw[black, ultra thick, dashed] (0.25\textwidth, 1.8) -- (0.25\textwidth, 0); 
        \end{tikzpicture}
        \caption*{\small 
        (3)
        \textbf{Prompt:} Reconstruct the missing \underline{eye}. \textbf{IoU:} 0.14}
    \end{subfigure}

    \begin{subfigure}[t]{0.45\textwidth}
        \centering
        \begin{tikzpicture}
            \node[anchor=south west, inner sep=0] (img) at (0,0)
                {\includegraphics[width=\textwidth]{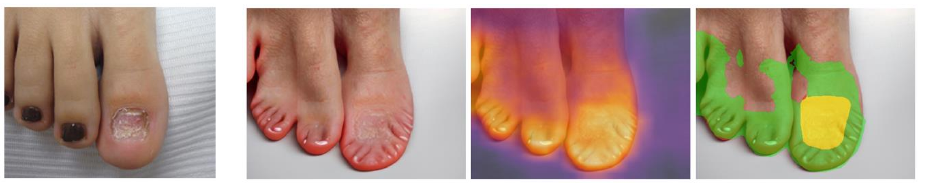}};
            \draw[black, ultra thick, dashed] (0.25\textwidth, 1.4) -- (0.25\textwidth, 0); 
        \end{tikzpicture}
        \caption*{\small 
        (4)
        \textbf{Prompt:} Reconstruct the big \underline{toenail}. \textbf{IoU:} 0.17}
    \end{subfigure}

    \begin{subfigure}[t]{0.45\textwidth}
        \centering
        \begin{tikzpicture}
            \node[anchor=south west, inner sep=0] (img) at (0,0)
                {\includegraphics[width=\textwidth]{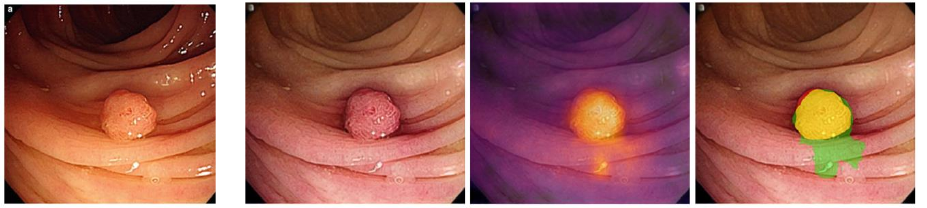}};
            \draw[black, ultra thick, dashed] (0.25\textwidth, 1.6) -- (0.25\textwidth, 0); 
        \end{tikzpicture}
        \caption*{\small 
        (5)
        \textbf{Prompt:} Remove the intestinal \underline{polyp}. \textbf{IoU:} 0.53}
    \end{subfigure}

    \begin{subfigure}[t]{0.45\textwidth}
        \centering
        \begin{tikzpicture}
            \node[anchor=south west, inner sep=0] (img) at (0,0)
                {\includegraphics[width=\textwidth]{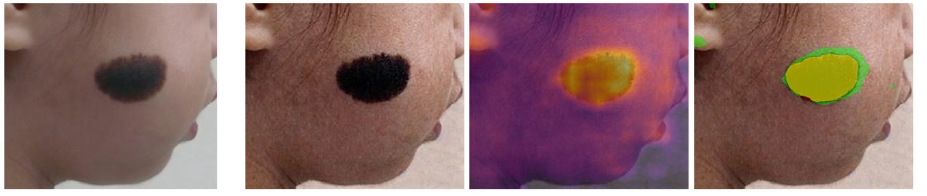}};
            \draw[black, ultra thick, dashed] (0.25\textwidth, 1.5) -- (0.25\textwidth, 0); 
        \end{tikzpicture}
        \caption*{\small 
        (6)
        \textbf{Prompt:} Remove \underline{birthmark}. \textbf{IoU:} 0.70}
    \end{subfigure}
    \caption{
    InstructPix2Pix cross-attention on key visual \underline{concept} versus ground-truth ROI. In the last column: red = ROI mask, green = attention, yellow = overlap.
    }

    \label{fig:composite_examples}
\end{figure}
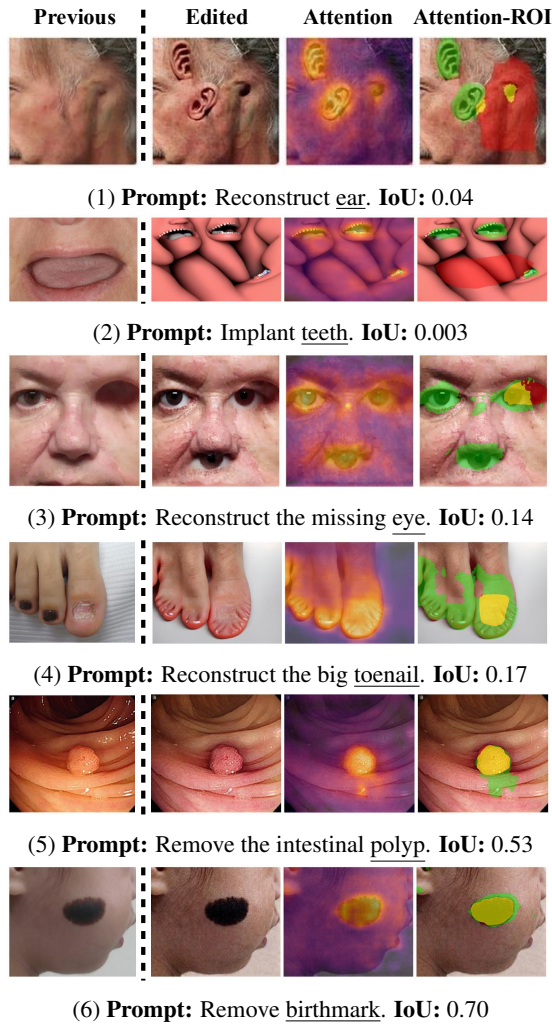
\begin{itemize}[noitemsep, topsep=0pt, leftmargin=*]
    \item \textbf{High IoU}: The model correctly localizes the target region but fails to apply the intended edit. This reflects partial spatial understanding with conservative or limited editing capability, often seen in concept removal tasks (e.g., samples 5–6 in Fig.~\ref{fig:composite_examples}).
   \item \textbf{Low IoU}: The model fails both to localize and to edit, attending to irrelevant regions. Such misalignment typically occurs in addition or reconstruction tasks, especially in anatomically complex areas (e.g., samples 1–4 in Fig.~\ref{fig:composite_examples}).
\end{itemize}
These patterns reveal fundamental limitations in current models’ spatial reasoning and medical concept grounding, underscoring the challenges of reliable medical image editing.

\vspace{-0.5em}
\section{Key Insights and Takeaways}

\vspace{-0.5em}
Our comprehensive evaluation across thirteen anatomical regions and multiple editing models yields four key findings:
\begin{itemize}[noitemsep, topsep=0pt, leftmargin=*]
    \item \textbf{Large multimodal models outperform open-source alternatives on complex medical edits.} Gemini 2 Flash consistently leads in EA, VQ, and FID, with the largest gaps in internal organ tasks (e.g., gastrointestinal tract, spine, teeth) where fine-grained structures and spatial reasoning are critical. Open-source models such as ICEdit remain competitive but lag in high-precision scenarios.

    \item \textbf{Region-specific challenges persist.} Models struggle with repetitive or occluded anatomy (e.g., hands, spine, teeth), while tasks involving superficial structures (e.g., skin blemishes, nose shape) are more reliably handled.

    \item \textbf{Fine-tuning aids domain adaptation; in-context learning shows limited generalization.} Fine-tuning models such as InstructPix2Pix improves medical editing performance despite data scarcity. However, large multimodal models struggle to transfer medical concepts through prompting-level in-context learning, highlighting generalization limits in clinical tasks.

    \item \textbf{Attention maps provide diagnostic insights.} IoU between attention heatmaps and ground-truth regions reveals failure patterns: high IoU with low EA indicates correct localization but failed execution (common in concept removal), while low IoU reflects poor spatial focus in addition or reconstruction tasks.
\end{itemize}

\FloatBarrier

\vspace{-0.5em}
\section{Related Works}
\vspace{-0.5em}

\textbf{Text-guided Image Editing} has advanced rapidly with diffusion-based models~\cite{ho2020ddpm, dhariwal2021diffusion, rombach2022latent, saharia2022imagen, nichol2021glide, ho2022cascaded, zhang2023controlnet, meng2021sdedit, ramesh2022dalle2, januspro2025}, enabling natural language-conditioned image synthesis~\cite{brooks2023instructpix2pix, kawar2022imagic, hertz2022prompt2prompt, mokady2022nulltext, choi2023customedit, ravi2023preditor, shen2022diffusionclip, shen2024swiftedit, shen2022sine, shen2023mdp,su2025thinkingimagesmultimodalreasoning}.  
Recent works~\cite{kawar2023imagictextbasedrealimage, brooks2023instructpix2pixlearningfollowimage, geng2023instructdiffusiongeneralistmodelinginterface, wasserman2025paintinpaintlearningadd, zhang2025incontexteditenablinginstructional, ge2025seedxmultimodalmodelsunified, google2024gemini2flash} further improve control, generalization, and real-world applicability. Medical image editing is an emerging application, supporting tasks such as disease progression simulation~\cite{taylor2019monitoring,puglisi2024enhancingspatiotemporaldiseaseprogression,cao2024medicalvideogenerationdisease, alaya2024mededitcounterfactualdiffusionbasedimage}, segmentation~\cite{feng2024enhancing,ma2025textdiffseg,dong2024tgediff,wu2023medsegdiffv2}, and synthetic data generation~\cite{cho2024medisyn,zhang2023diffboost,zhang2024textguided,kidder2024advanced,li2024synthetic}.

\noindent \textbf{Benchmarking Text-Guided Editing} is key to evaluating model performance, but standard metrics such as FID~\cite{heusel2018ganstrainedtimescaleupdate}, CLIP Score~\cite{hessel2022clipscorereferencefreeevaluationmetric}, PSNR~\cite{korhonen2012psnr}, SSIM~\cite{wang2004ssim}, and LPIPS~\cite{zhang2018lpips} fail to capture fine-grained edit quality and semantic intent. Recent benchmarks~\cite{ma2024i2ebenchcomprehensivebenchmarkinstructionbased, basu2023editvalbenchmarkingdiffusionbased, wang2023imageneditoreditbenchadvancing, ju2023directinversionboostingdiffusionbased} improve on this with human-aligned dimensions and diverse edit types. Multimodal Large Language Models (MLLMs), such as GPT-4o~\cite{openai2024gpt4o} and Gemini 2.5 Pro~\cite{google2025gemini25pro}, have emerged as strong evaluators for visual and semantic alignment, enabling interpretable and human-aligned assessment frameworks~\cite{li2024m3agiqa, jin2024adaptive, zhang2024grounding, liu2024comprehensive, chen2023xiqeexplainableimagequality}.  However, existing work largely targets general-domain imagery, overlooking domain-specific needs such as medical editing. To bridge this gap, we propose \textbf{MedEBench}, the first benchmark for text-guided medical image editing with a clinically grounded evaluation framework.

\vspace{-0.5em}
\section{Conclusion}

\vspace{-0.5em}
We introduce \textbf{MedEBench}, the first benchmark specifically designed for text-guided medical image editing. It includes 1,182 real-world examples spanning 13 anatomical regions and 70 clinically meaningful tasks. Each case is annotated with region-of-interest masks and expert-authored change descriptions, enabling multi-faceted evaluation across \textit{Editing Accuracy}, \textit{Context Preservation}, and \textit{Visual Quality}.

We benchmark seven state-of-the-art models and find persistent challenges, particularly in editing internal organs and anatomically complex regions. While Gemini 2 Flash achieves the highest overall performance, ICEdit stands out as the strongest open-source alternative. Fine-tuning diffusion-based, text-guided editing models like InstructPix2Pix leads to substantial performance gains, even in low-data settings, highlighting the critical role of domain adaptation. In contrast, large multimodal models often fail to generalize medical concepts through prompt-based in-context learning, pointing to core limitations in their clinical transferability. Further analysis of cross-attention patterns reveals consistent gaps in spatial and anatomical grounding, indicating a mismatch between model focus and human-intended edits.

Looking ahead, our findings emphasize the need for anatomy-aware architectures and medically aligned supervision to improve reliability and safety in clinical image editing tasks. We hope \textbf{MedEBench} will serve as a catalyst for developing robust, transparent, and domain-adapted generative tools for medical applications.



\section*{Limitations}
While MedEBench provides a comprehensive benchmark for text-guided medical image editing, several limitations remain. First, the benchmark focuses exclusively on editing tasks that correspond to real-world surgical or clinical operations. As a result, it does not cover more speculative or exploratory editing tasks that may be of interest for rare disease modeling, or synthetic data augmentation. This focus ensures clinical relevance but limits the diversity of task types included. Nevertheless, extending the benchmark with resources such as GARD~\cite{gard_ncbi}, Orphanet~\cite{rath2009orphanet}, RaDaR~\cite{radar_ukkidney}, and MONAI~\cite{monai_2025} represents a promising future direction.

Second, current text-to-mask models, such as Grounded-SAM, face significant challenges when applied to medical images. Specifically, these models struggle to accurately generate region-of-interest (ROI) masks for anatomical structures without clear boundaries or with ambiguous visual features. In such cases, automatic mask generation often fails, necessitating human intervention to ensure anatomical precision.

Finally, while MedEBench primarily sources image pairs from Creative Commons Search and open-access publications, all data have been carefully curated to ensure clinical plausibility and anatomical relevance. Any identifiable features are anonymized through masking or blurring, but reliance on public data may constrain dataset diversity compared to clinical repositories.

\section*{Acknowledgments}
We would like to thank Hongrui Liu from the School of Basic Medical Sciences, Peking University, and Siman Song from the School of Clinical Medicine at China Medical University for their domain expertise in medical image analysis.
\section*{Ethical considerations}

MedEBench is constructed using publicly available and de-identified medical images.  To ensure transparency and reproducibility, the original URLs of all raw images are provided as part of the dataset release. No personally identifiable information (PII) or sensitive patient data is included. The benchmark strictly serves research purposes and does not support diagnostic, therapeutic, or clinical decision-making applications. Although editing tasks are designed to correspond to real-world surgical procedures, the benchmark does not promote automated clinical image generation without human expert oversight, and any misuse for unauthorized medical applications is strongly discouraged. In cases where text-to-mask models such as Grounded-SAM fail to generate reliable masks, human annotators with medical expertise are employed to ensure anatomical accuracy under proper consent and contractual agreements. Recognizing the potential for dataset-induced biases, we strive for diversity across anatomical regions and clinical tasks, while acknowledging that further work is needed to address bias mitigation in synthetic image generation. The benchmark, models, and code are released solely for academic and non-commercial use under an appropriate open-source license, with a strict reminder that MedEBench is not a substitute for clinical judgment or professional medical practice.

Despite the intended academic use of MedEBench, there are inherent risks associated with the misuse of medical image editing models. One primary concern is the generation of misleading or fabricated clinical images that could potentially be used in malicious contexts, such as misinformation, fraudulent clinical documentation, or unauthorized patient record manipulation. Additionally, the use of generative models trained on limited or biased datasets may inadvertently reinforce existing healthcare disparities if applied to populations not well represented in the benchmark. Another risk lies in the potential over-reliance on automated editing tools without sufficient medical expertise, which could lead to clinically inaccurate or unsafe modifications. To mitigate these risks, we emphasize responsible usage under expert supervision, explicit academic licensing terms, and continuous community oversight.

\bibliography{custom, anthology}     
\clearpage
\appendix
\onecolumn
\FloatBarrier

\section{Introduction}
\begin{figure*}[htbp]
\centering
\begin{tikzpicture}
    \node[anchor=south west, inner sep=0] (image) at (0,0)
    {\includegraphics[width=0.95\textwidth]{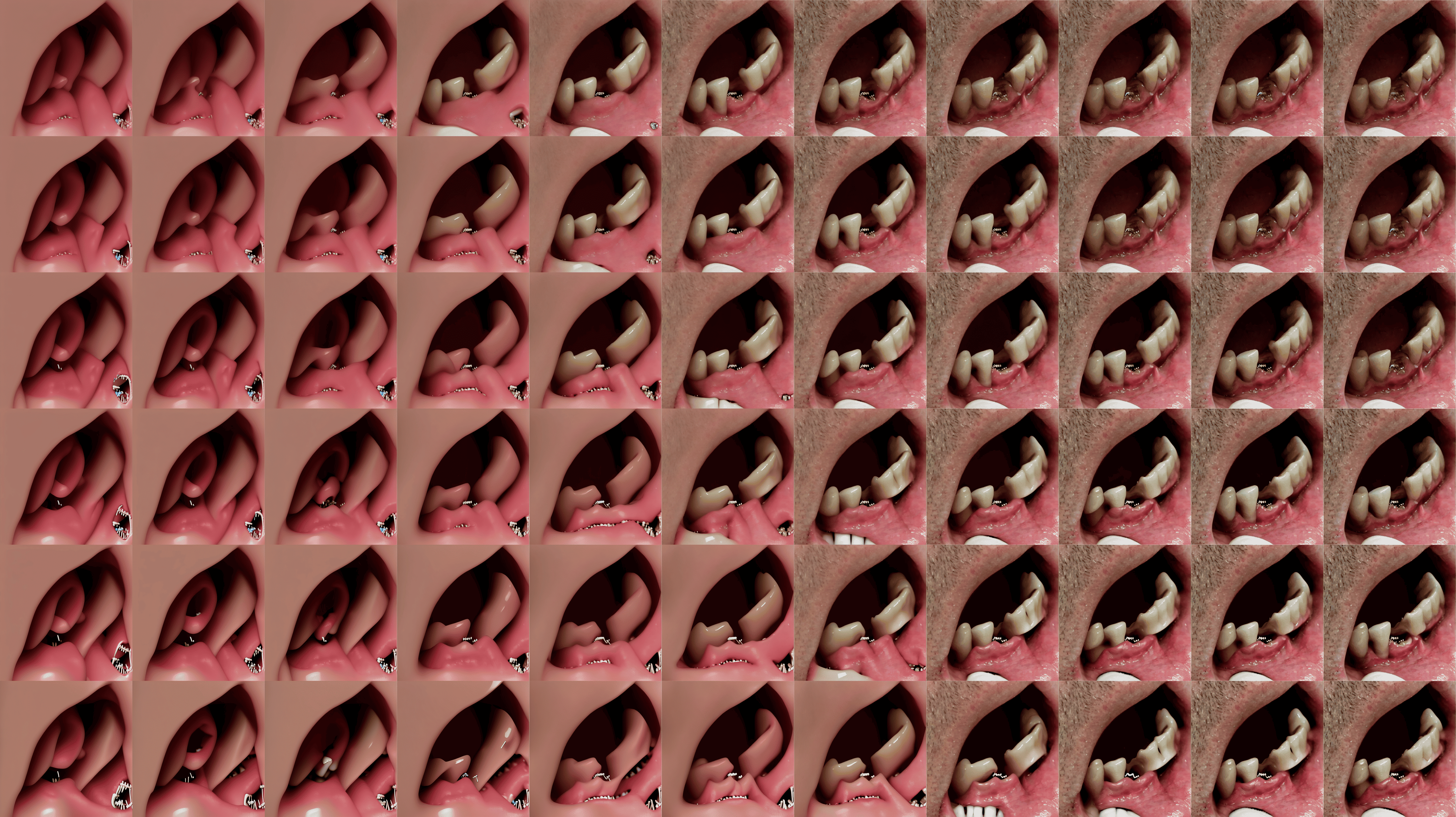}};
    \draw[->, thick] (0,0) -- (0,9.3) node[right] {\textbf{Text Guidance Scale}};
    \draw[->, thick] (0,0) -- (16,0) node[below,xshift=-2cm] {\textbf{Image Guidance Scale}};
\end{tikzpicture}
\caption{Visualization of the effect of varying text and image guidance scales for the task of adding a missing tooth. Despite adjustments, the desired edit could not be achieved by InstructPix2Pix.}
\label{fig:labeled_axes}
\end{figure*}

Figure~\ref{fig:labeled_axes} illustrates the limitations of controlling text and image guidance scales in diffusion-based image editing. Specifically, for the task of “adding a missing tooth,” varying these scales fails to yield a satisfactory result when using the InstructPix2Pix model. While increasing the text guidance scale emphasizes the semantic prompt, and higher image guidance preserves visual fidelity to the original image, neither direction successfully produces the desired anatomical modification. This outcome suggests that simply tuning global guidance weights is insufficient for achieving fine-grained, localized edits in medical or detail-critical domains. As shown in Figure~\ref{fig:labeled_axes}, the generated outputs either omit the new tooth entirely or introduce unnatural artifacts, underscoring the need for more controllable and spatially-aware editing approaches.

\section{Details of MedEBench Dataset}

Most images were collected using Google Search with Creative Commons license filters. Some additional images were taken from figures in publicly available open-access papers published under the CC BY 4.0 license, including journals such as MDPI, Frontiers, PLOS, eLife, BMC Bioinformatics, BMC Medicine and Hindawi. To promote transparency, we provide source URLs for all images whenever possible.

Table~\ref{tab:organ_task_sample_count} summarizes the distribution of organ-related tasks included in MedEBench, focusing on tasks with more than five samples.

\begin{figure*}[h!]
\centering
\begin{minipage}[t]{0.48\textwidth}
\vspace*{-21\baselineskip} 
\begin{lstlisting}[language=json]
{
  "Id": 1,
  "Organ": "Teeth",
  "Task": "Implant or add missing teeth.",
  "Prompt": "Add a tooth in the missing area.",
  "Rephrased_prompt": "Place a tooth where one is absent",
  "Detailed_description": "The image modality is intraoral photography; the addition of a tooth in the lower dental arch was performed to fill the gap, resulting in a complete and continuous row of teeth with natural alignment and spacing.",
  "Previous_image": "editing/previous/1.png",
  "GroundTruth_image": "editing/changed/1.png",
  "ROI_mask": "editing/previous_mask/1.png",
  "url": "https://dentistpeshawar.pk/wp-content/uploads/2024/07/extraction-and-implant-same-day.jpg"
}
\end{lstlisting}
\captionof{figure}{Example metadata entry in MedEBench.}
\label{fig:metadata_example}
\end{minipage}%
\hfill
\begin{minipage}[t]{0.48\textwidth}
\centering
\includegraphics[width=\linewidth]{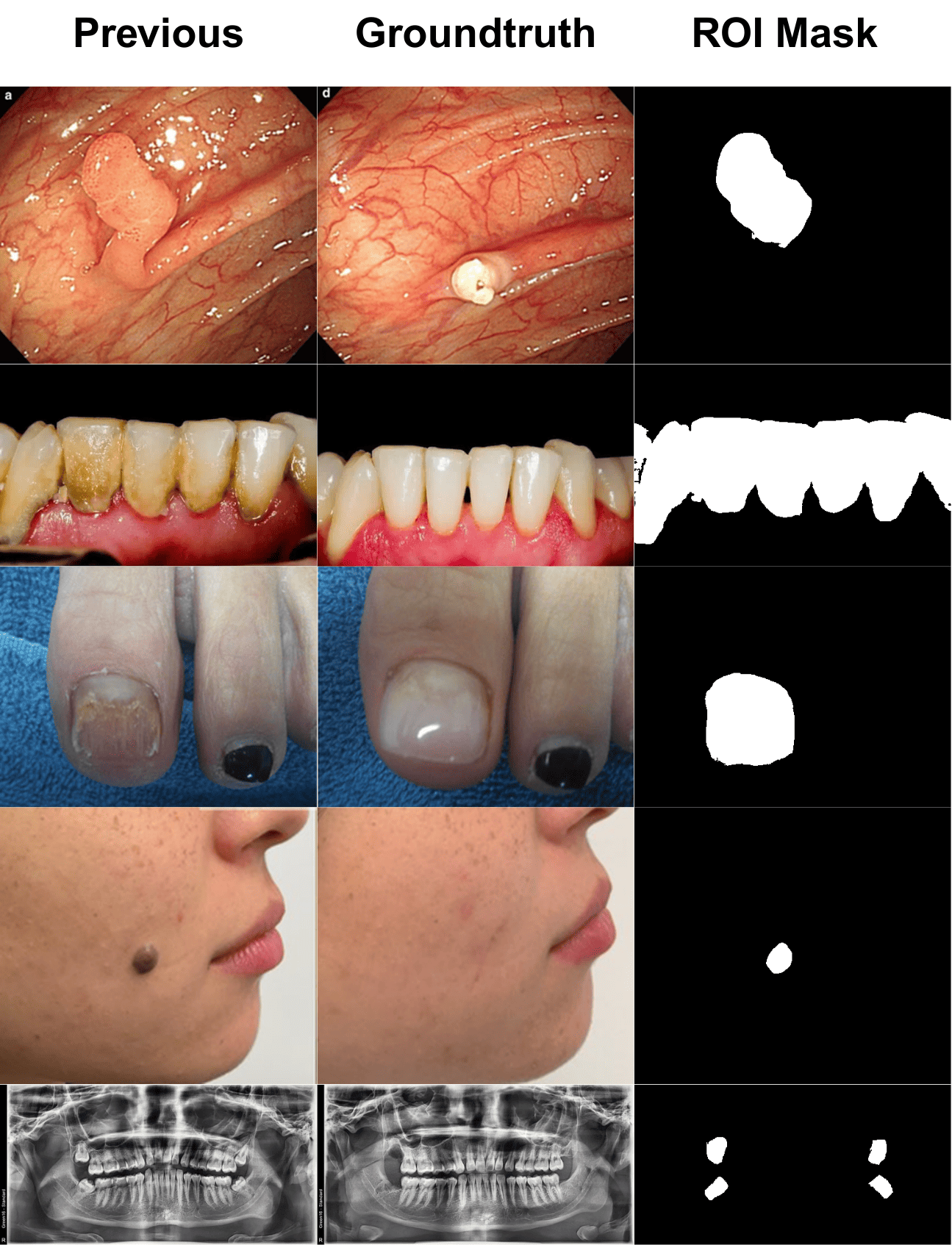}
\captionof{figure}{Example groups of previous groundtruth images and ROI mask.}
\label{fig:appendix_dataset_example}
\end{minipage}
\end{figure*}

Each sample in the MedEBench dataset is defined by a structured metadata entry that includes the editing prompt, a description of the expected change, a pair of previous and ground truth images, and an annotated ROI mask on the previous image. The URL of the source image is also provided. Examples are shown in Fig.~\ref{fig:metadata_example} and Fig.~\ref{fig:appendix_dataset_example}.

\begin{table}[h!]
\begin{tabular}{l l c}
\toprule
\textbf{Organ} & \textbf{Task} & \textbf{Sample Count} \\
\midrule
\multirow{2}{*}{Ears} & Reconstruct ear. & 31 \\
  & Reconstruct auricle. & 11 \\
\midrule
\multirow{8}{*}{Eyes} & Reconstruct missing or injured eye. & 35 \\
  & Reconstruct eyelid or lower eyelid skin. & 29 \\
  & Create double eyelids. & 29 \\
  & Remove eyelid styes (chalazion). & 14 \\
  & Correct lower eyelid ectropion. & 10 \\
  & Remove conjunctival nevus. & 8 \\
  & Remove eye bags. & 8 \\
  & Remove eyelid xanthalasma. & 6 \\
\midrule
\multirow{7}{*}{Face} & Reduce facial wrinkles. & 25 \\
  & Remove or revise facial scars. & 21 \\
  & Reconstruct cheek or forehead skin. & 17 \\
  & Remove facial redness. & 17 \\
  & Remove excess fat from neck. & 15 \\
  & Lift neck and face by tightening skin. & 10 \\
  & Remove facial acne, bumps, or cysts. & 7 \\
\midrule
\multirow{1}{*}{Feet} & Repair or reconstruct toenail. & 30 \\
\midrule
\multirow{1}{*}{Gastrointestinal Tract} & Remove intestinal polyps or adenomas. & 15 \\
\midrule
\multirow{3}{*}{Hair} & Make hair thicker. & 150 \\
  & Make beard thicker. & 13 \\
  & Make eyebrow thicker. & 8 \\
\midrule
\multirow{2}{*}{Hands} & Complete missing finger or fingernail. & 18 \\
  & Improve hand appearance by injectable filler. & 7 \\
\midrule
\multirow{2}{*}{Mouth} & Reconstruct damaged lip or lip skin. & 20 \\
  & Perform lip augmentation. & 11 \\
\midrule
\multirow{3}{*}{Nose} & Reconstruct or repair nose skin. & 107 \\
  & Remove or fade scar on nose. & 23 \\
  & Remove or beautify rhinophyma. & 16 \\
\midrule
\multirow{6}{*}{Skin} & Remove moles, nevi, or black marks. & 136 \\
  & Reconstruct damaged skin or scalp. & 48 \\
  & Remove varicose veins. & 36 \\
  & Remove brown spots or pigmentation. & 31 \\
  & Remove or fade scars. & 20 \\
  & Remove black birthmarks to even skin tone. & 14 \\
\midrule
\multirow{2}{*}{Spine and Bones} & Correct spine alignment. & 10 \\
  & Fix fractures with screws or splints. & 6 \\
\midrule
\multirow{4}{*}{Teeth} & Remove wisdom teeth. & 31 \\
  & Remove stains, tartar, or plaque from teeth. & 22 \\
  & Implant or add missing teeth. & 13 \\
  & Repair or restore damaged teeth. & 13 \\
\midrule
\multirow{1}{*}{Torso} & Perform body liposuction and skin tightening. & 36 \\
\bottomrule
\end{tabular}
\centering
\caption{Tasks with More Than 5 Samples per Organ}
\label{tab:organ_task_sample_count}
\end{table}

\section{Baseline Models Implementation Detail}
We summarize the inference-time configurations for each baseline model as follows:
\begin{itemize}[noitemsep,topsep=0pt,leftmargin=1.5em]
    \item \textbf{InstructPix2Pix}: \texttt{DIFFUSION\_STEPS} = 50, \texttt{image\_guidance\_scale} $\in$ \{1.55, 1.6, 1.65\}, \texttt{guidance\_scale} $\in$ \{7.5, 7.6\}.
    \item \textbf{Imagic}: \texttt{num\_inference\_steps} = 50, \texttt{alpha} $\in$ \{1.3, 1.35\}, \texttt{guidance\_scale} $\in$ \{7.5, 7.6\}.    
    \item \textbf{InstructDiffusion}: \texttt{cfg\_text} = 5.0, \texttt{cfg\_image} = 1.25.   
    \item \textbf{Paint-by-Inpaint}: \texttt{DIFFUSION\_STEPS} = 50, \texttt{image\_guidance\_scale} = 1.7, \texttt{guidance\_scale} = 7.0.   
    \item \textbf{ICEdit}: \texttt{num\_inference\_steps} = 28, \texttt{guidance\_scale} = 50.    
    \item \textbf{SEED-X}: \texttt{num\_inference\_steps} = 50, instruction format: \texttt{"Edit this image: "} + editing prompt.   
    \item \textbf{Gemini 2.0 Flash}: prompt format: \texttt{"You are good at image editing. Here is the image editing instruction: "} + editing instruction.
\end{itemize}

\section{Human Evaluation and Validation Criteria}
\subsection{Validating Change of Difference Descriptions}
1. \textbf{Structured Prompting}:  
   Before rephrasing the output into free-form descriptions, we guided GPT-4o’s generation using a structured template: "Edit this {modality} to simulate the {action} of the {entity} from the {region}, while preserving the {anatomical context}.
   This structured approach constrained the model to produce outputs that were more consistent, structured, and clinically relevant, ensuring that the generated descriptions adhered to medical terminology.

2. \textbf{Independent Human Validation}:  
   Each generated description was independently reviewed by two senior medical students. Their task was to verify that every component of the description (modality, action, entity, region, and anatomical context) accurately reflected the visual changes in the corresponding image pair and adhered to standard medical terminology. Importantly, the students were not involved in the subsequent image editing quality evaluations, ensuring an unbiased assessment. 

   \textbf{Validation Results}:  
   243 out of 260 descriptions (93.5\%) were confirmed by both reviewers as fully accurate, with no modifications required. The remaining 17 descriptions (6.5\%) contained minor inaccuracies or imprecise phrasing and were manually corrected by the reviewers before being used.

This human-in-the-loop methodology effectively mitigates the risks associated with AI generation, ensuring that the descriptions are reliable and of high fidelity.

\subsection{Human Evaluation Study}
\begin{figure*}[h!]
\centering
\includegraphics[width=\textwidth]{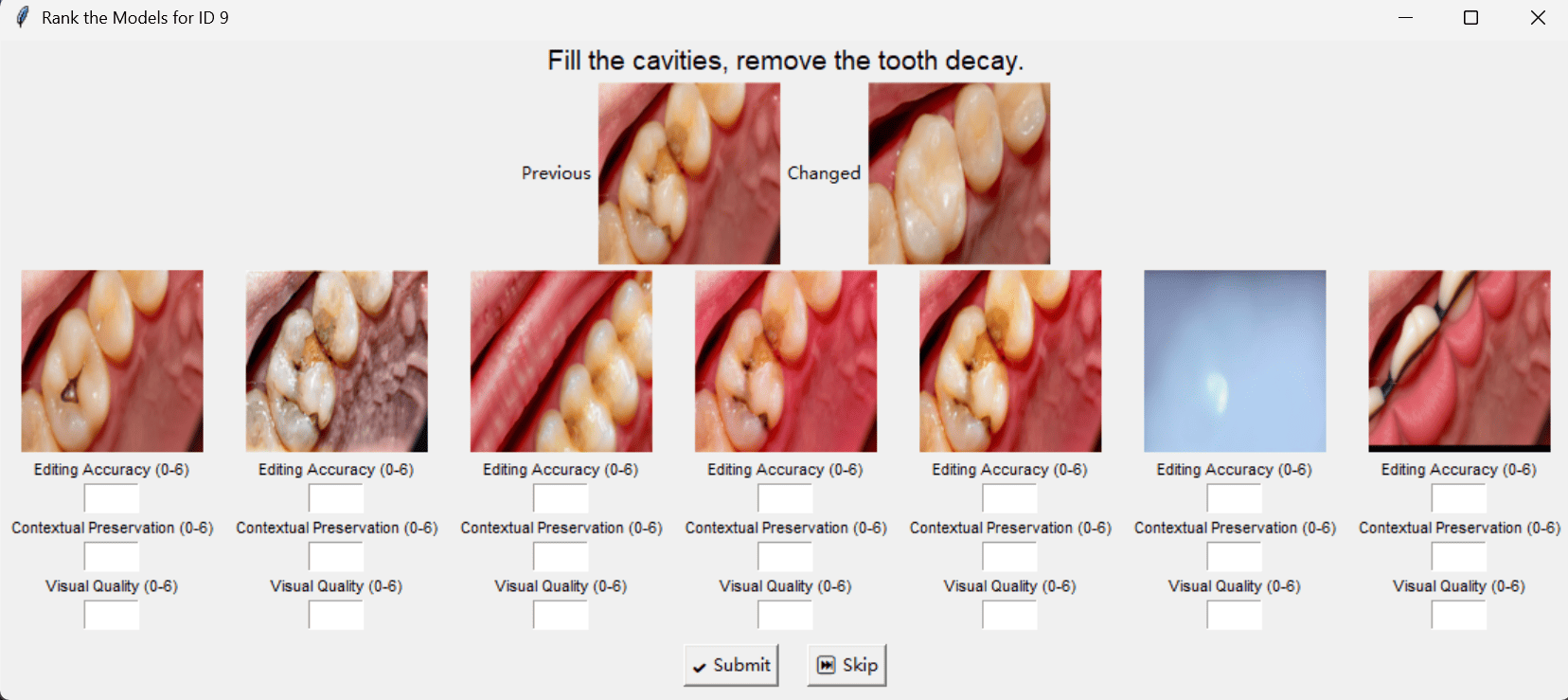}
\caption{
\textbf{Human Evaluation Panel.} Annotators were instructed to rank edited images based on three dimensions: 
\textbf{Editing Accuracy (EA)} — whether the intended anatomical modification has been correctly and plausibly applied as described in the prompt; 
\textbf{Contextual Preservation (CP)} — whether unedited regions maintain their original anatomical structure without unintended alterations; 
and \textbf{Visual Quality (VQ)} — the overall perceptual quality of the image, including the seamlessness of edits, absence of artifacts, clarity, and realistic color fidelity.
}
\label{fig:human_eval_panel}
\end{figure*}

To validate the effectiveness of our automated metrics, we conducted a human evaluation study using a relative ranking protocol. Two expert annotators independently assessed a subset of edited images across three dimensions—\textit{Editing Accuracy} (EA), \textit{Contextual Preservation} (CP), and \textit{Visual Quality} (VQ)—which align with the dimensions used in our automated evaluation framework. The evaluation interface and annotator instructions are illustrated in Fig.~\ref{fig:human_eval_panel}. For each organ $o \in \mathcal{O}$, we randomly sampled 20 editing tasks, resulting in a total of 260 edited image samples. Outputs from multiple models were collected using identical inputs and prompts to ensure fair comparison. Two annotators with bachelor’s degrees were hired from a crowdsourcing platform, compensated at a rate of 4.2 HKD per sample. They were instructed to rank the model outputs for each sample along all three dimensions. The inter-annotator agreement, measured by the Spearman rank correlation coefficient, reached 0.91, indicating excellent consistency and high annotation reliability.

\section{Automated Evaluation Pipeline}

\begin{figure*}[h!]
\centering
\includegraphics[width=\textwidth]{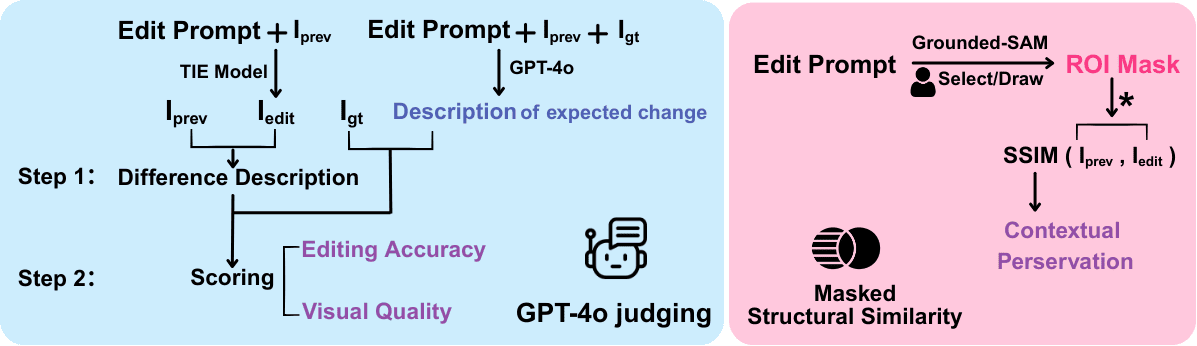}
\caption{Automatic Evaluation Pipeline in detail.}
\label{fig:Detailed_eval_pipe}
\end{figure*}

Figure~\ref{fig:Detailed_eval_pipe} provides an overview of our automatic evaluation pipeline, which consists of two complementary components: GPT-4o-based judgment and masked SSIM measurement.

On the left side of the figure, we illustrate a two-step process for MLLM-based evaluation. In Step 1, GPT-4o compare the original image ($I_{\text{prev}}$) and the edited image ($I_{\text{edit}}$) from a TIE model to generate a description of the actual visual difference. Simultaneously, we also use GPT-4o to produce a reference description of the expected change, given the edit prompt and the pair $(I_{\text{prev}}, I_{\text{gt}})$, where $I_{\text{gt}}$ is the ideal target image. In Step 2, GPT-4o compares the actual and expected change descriptions and produces two scores: \textit{Editing Accuracy}, which reflects how well the edit aligns with the prompt, and \textit{Visual Quality}, which assesses the perceptual realism and consistency of the output.

On the right side of the figure, we assess \textit{Contextual Preservation} using masked structural similarity. An ROI mask is generated using a prompt-guided method such as Grounded-SAM or through manual selection. We then compute SSIM between $I_{\text{prev}}$ and $I_{\text{edit}}$, restricted to the masked region's complement. This quantifies how much of the image outside the intended region was unnecessarily altered.

\subsection{An Example to Demonstrate the Evaluation Process}

\begin{figure*}[h!]
\centering
\includegraphics[width=\textwidth]{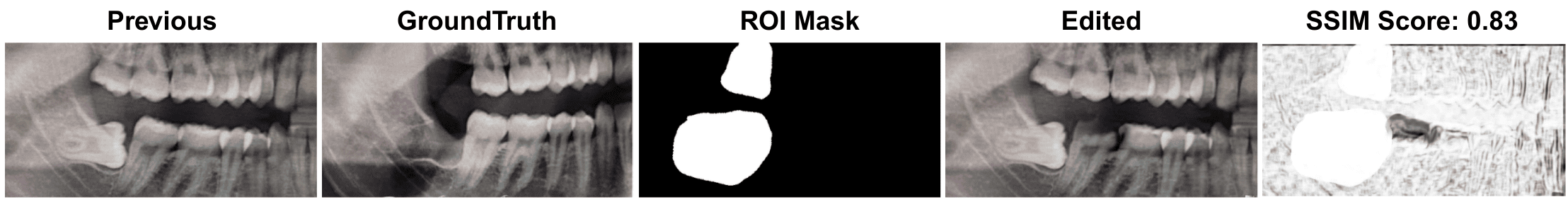}
\caption{Evaluation example for contextual preservation. The contextual preservation score (masked SSIM) is 0.83.}
\label{fig:An_example}
\end{figure*}

Figure~\ref{fig:An_example} presents a representative evaluation result for a sample on Contextual Preservation. The corresponding GPT-4o-based evaluation of editing accuracy and visual quality is shown below.

\textbf{Edit Prompt:} ``Remove the wisdom teeth.''

\textbf{Description of Expected Change:} ``Panoramic dental radiograph: Wisdom teeth in the upper and lower jaws were removed, resulting in a post-surgical appearance with space in the alveolar bone where the teeth previously resided.'' 

\vspace{0.5em}
\newpage
\begin{tcolorbox}[
  title=GPT-4o Evaluation Output for the Sample in Fig.~\ref{fig:An_example},
  colback=gray!3,
  colframe=gray!50,
  breakable,
  enhanced jigsaw,
  sharp corners=south,
  fonttitle=\bfseries,
]

\textbf{Visual Difference Description}

\begin{itemize}
  \item \textbf{Lower Wisdom Tooth (Mandibular Third Molar):}
    \begin{itemize}
      \item The impacted lower left wisdom tooth remains clearly visible in the edited image.
      \item There is no evidence of surgical removal or space in the alveolar bone.
      \item No post-surgical bone contour or socket is visible — it is identical to the input.
    \end{itemize}
    
  \item \textbf{Upper Wisdom Tooth (Maxillary Third Molar):}
    \begin{itemize}
      \item The upper left third molar is also present and unchanged in the edited image.
      \item Again, no extraction space or bone remodeling appears.
    \end{itemize}
    
  \item \textbf{No Visible Surgical Artifacts:}
    \begin{itemize}
      \item There are no additions or edits indicating removal, such as bone healing patterns, shadow variations, or softening around the third molar regions.
    \end{itemize}

  \item \textbf{Image Tone and Contrast:}
    \begin{itemize}
      \item The edited image has slightly more contrast and minor brightness variation.
      \item These changes appear incidental and are not related to anatomical edits.
    \end{itemize}
\end{itemize}

\vspace{0.5em}
\textbf{Step 2: Evaluation}

\begin{itemize}
  \item \textbf{Editing Accuracy: 0/10}
    \begin{itemize}
      \item Neither the upper nor lower wisdom teeth were removed.
      \item The anatomical regions remain untouched.
      \item No attempt at post-extraction rendering (e.g., healed socket) is visible.
      \item The edit does not align with the prompt at all.
    \end{itemize}
    
  \item \textbf{Visual Quality: 8/10}
    \begin{itemize}
      \item The edited image retains good clarity and sharpness.
      \item The contrast is slightly increased but does not degrade quality.
      \item No artifacts, glitches, or unrealistic elements are introduced — the image remains diagnostically useful.
    \end{itemize}
\end{itemize}

\end{tcolorbox}

\subsection{Additional Editing Results}
Table~\ref{tab:editing_table} provides visual comparisons of editing results from 7 baseline models. Each example shows the input, ground truth, and outputs from various methods, with corresponding scores for EA, CP, and VQ.
\begin{table*}[htbp]
    \centering
    \renewcommand{\arraystretch}{1.2}
    \setlength{\tabcolsep}{4pt}
    \begin{tabular}{ccc@{}c@{}c@{}c@{}c@{}c@{}c@{}}
    \scriptsize\textbf{Previous} & \scriptsize\textbf{Truth} & \scriptsize\textbf{Gemini2} & \scriptsize\textbf{SeedX} & \scriptsize\textbf{Imagic} & \scriptsize\textbf{IP2P} & \scriptsize\textbf{InstructDiff.} & \scriptsize\textbf{PaintByInpaint} & \scriptsize\textbf{ICEdit} \\
    
    \midrule
    \includegraphics[width=0.1\textwidth]{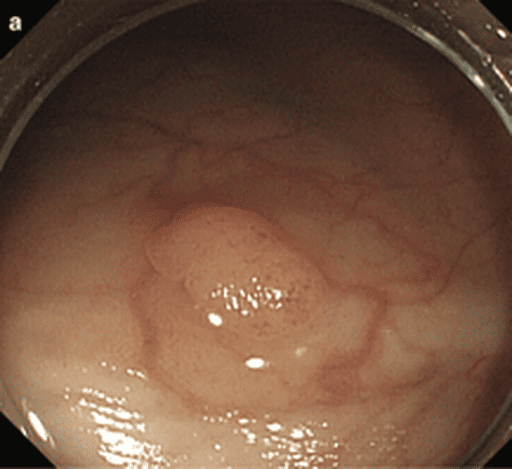} & \includegraphics[width=0.1\textwidth]{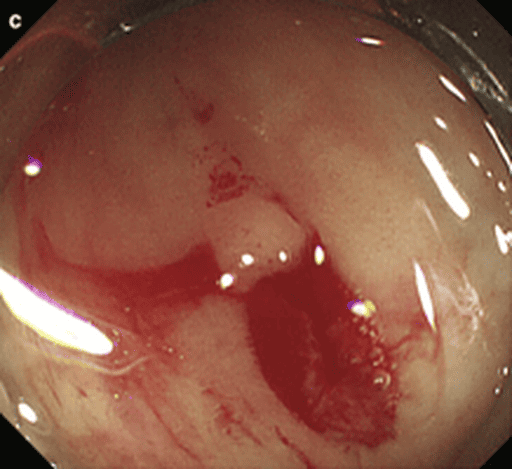} & \includegraphics[width=0.1\textwidth]{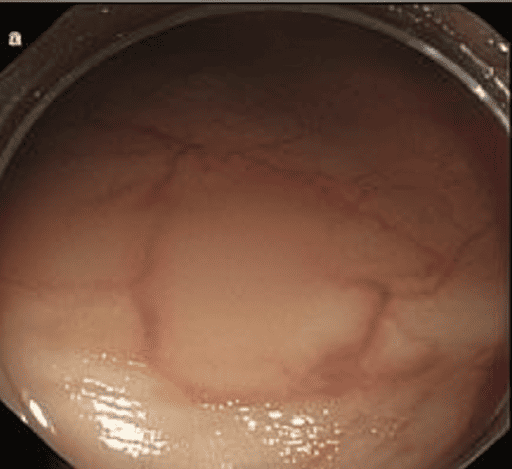} & \includegraphics[width=0.1\textwidth]{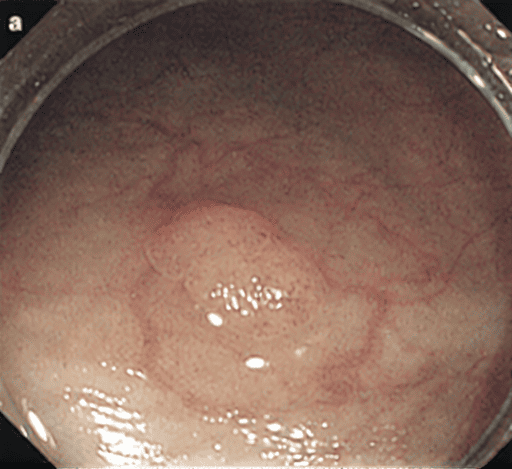} & \includegraphics[width=0.1\textwidth]{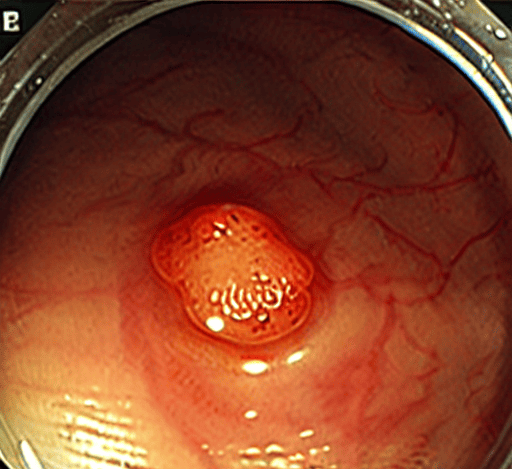} & \includegraphics[width=0.1\textwidth]{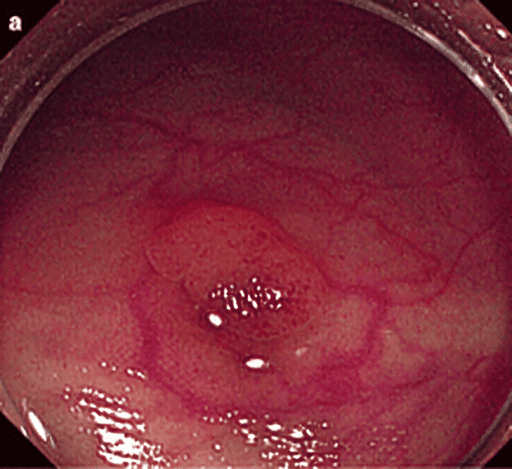} & \includegraphics[width=0.1\textwidth]{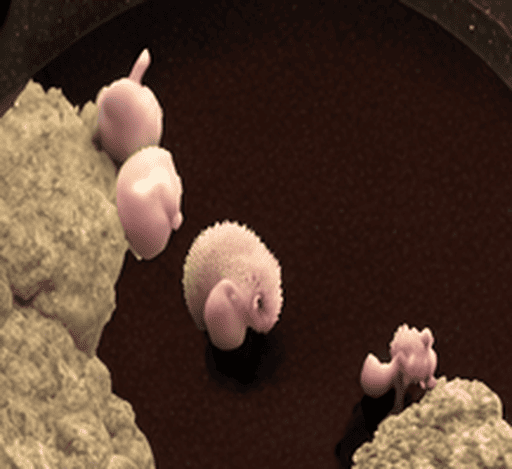} & \includegraphics[width=0.1\textwidth]{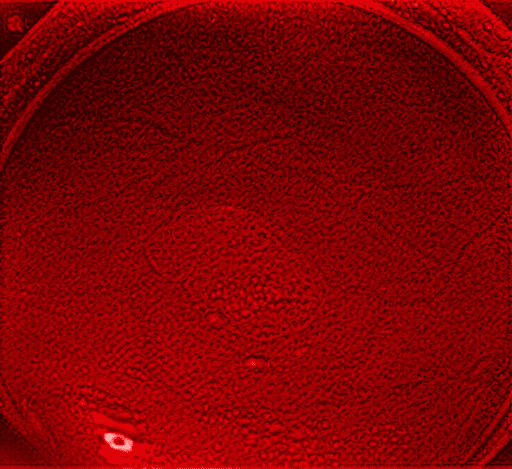} & \includegraphics[width=0.1\textwidth]{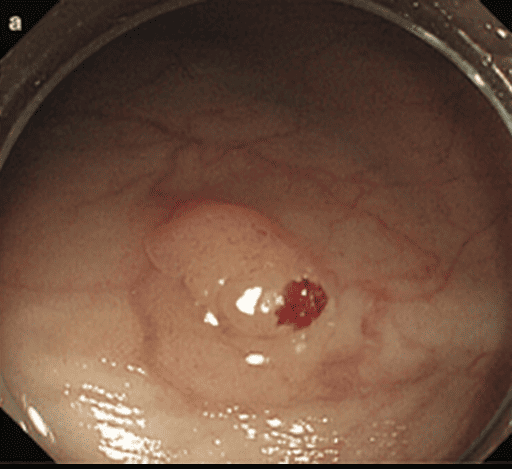} \\
    \multicolumn{2}{c}{\small Remove the diminutive polyp.} & \small 0.8/0.8/0.9 & \small 0.1/1.0/1.0 & \small 0.9/0.9/0.9 & \small 0.6/0.7/1.0 & \small 0.0/0.3/0.9 & \small 0.2/0.2/0.8 & \small 0.9/0.9/1.0 \\
    \midrule
    \includegraphics[width=0.1\textwidth]{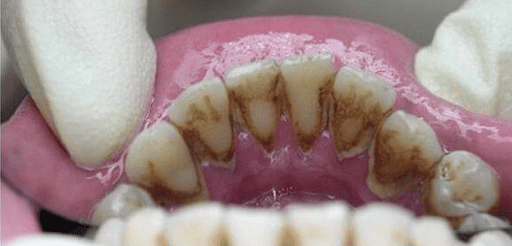} & \includegraphics[width=0.1\textwidth]{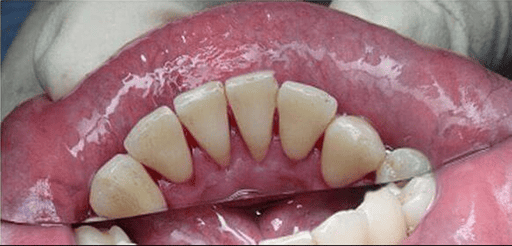} & \includegraphics[width=0.1\textwidth]{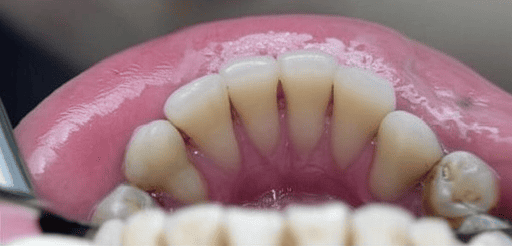} & \includegraphics[width=0.1\textwidth]{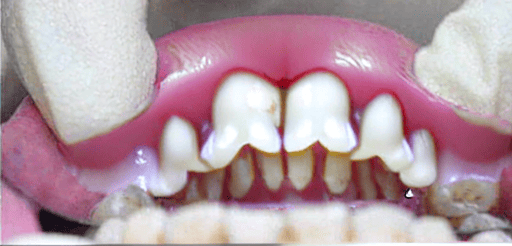} & \includegraphics[width=0.1\textwidth]{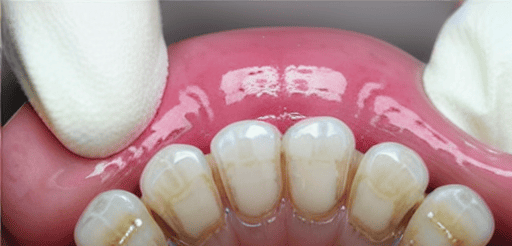} & \includegraphics[width=0.1\textwidth]{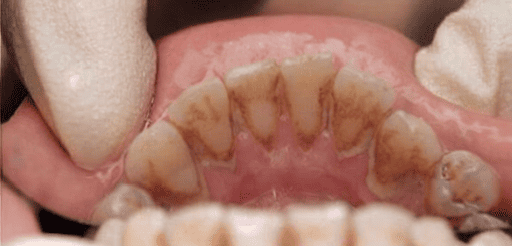} & \includegraphics[width=0.1\textwidth]{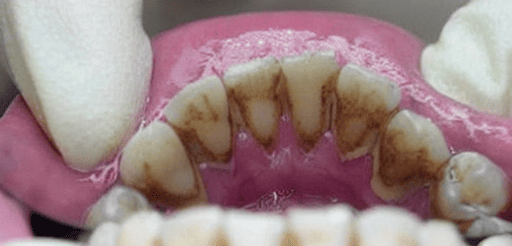} & \includegraphics[width=0.1\textwidth]{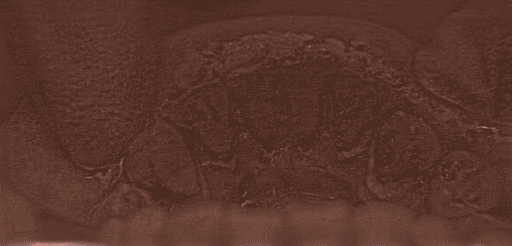} & \includegraphics[width=0.1\textwidth]{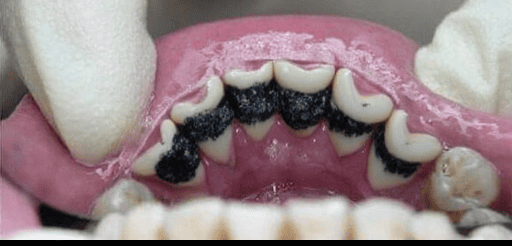} \\
    \multicolumn{2}{c}{\small Remove the dental black stains.} & \small 0.9/0.8/0.8 & \small 0.7/0.5/0.9 & \small 0.9/0.8/0.7 & \small 0.5/0.5/0.9 & \small 0.5/0.7/0.7 & \small 0.2/0.1/0.6 & \small 0.0/0.5/0.8 \\
    \midrule
    \includegraphics[width=0.1\textwidth]{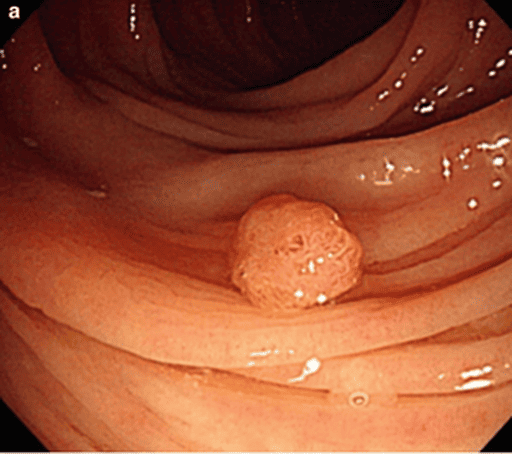} & \includegraphics[width=0.1\textwidth]{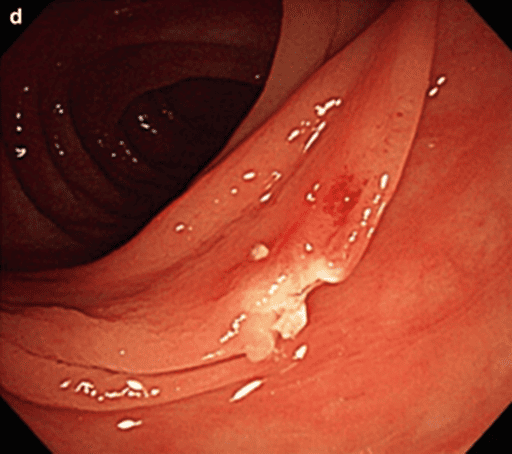} & \includegraphics[width=0.1\textwidth]{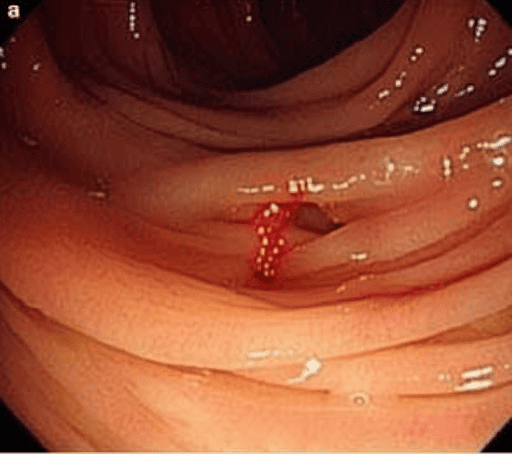} & \includegraphics[width=0.1\textwidth]{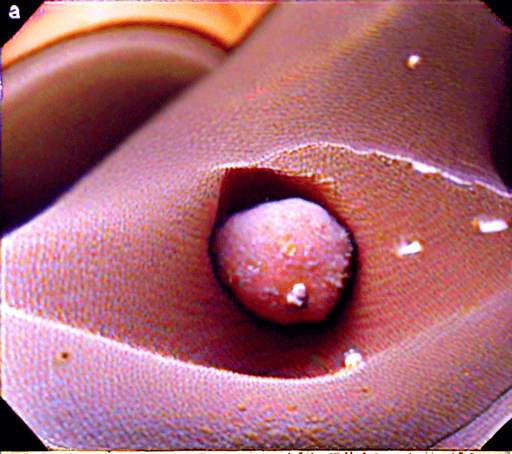} & \includegraphics[width=0.1\textwidth]{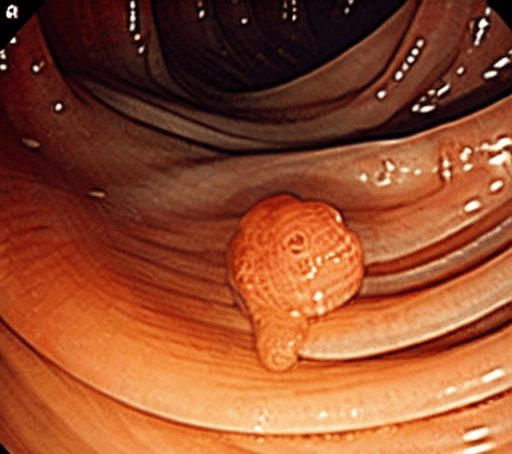} & \includegraphics[width=0.1\textwidth]{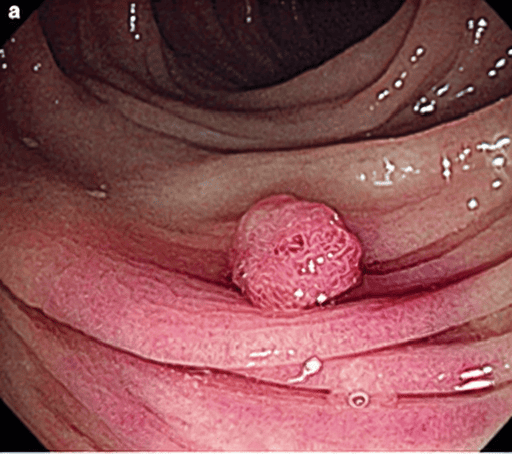} & \includegraphics[width=0.1\textwidth]{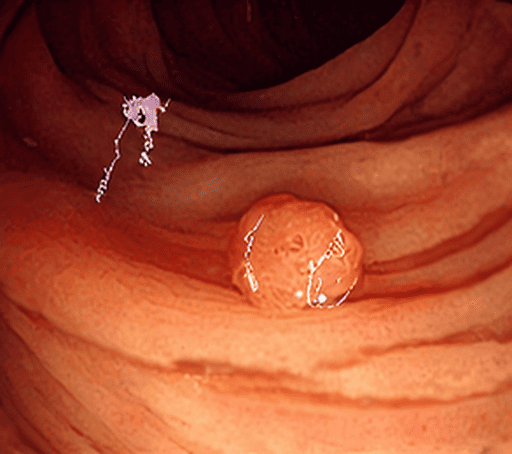} & \includegraphics[width=0.1\textwidth]{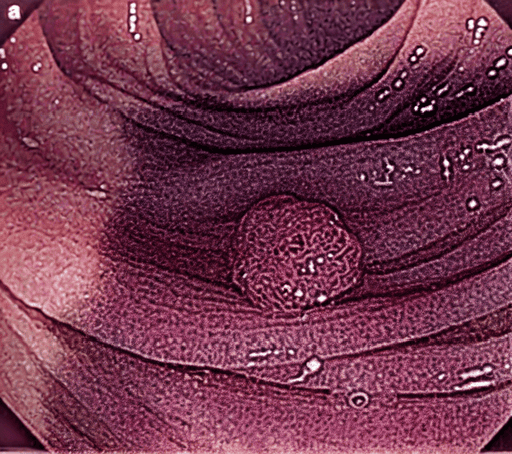} & \includegraphics[width=0.1\textwidth]{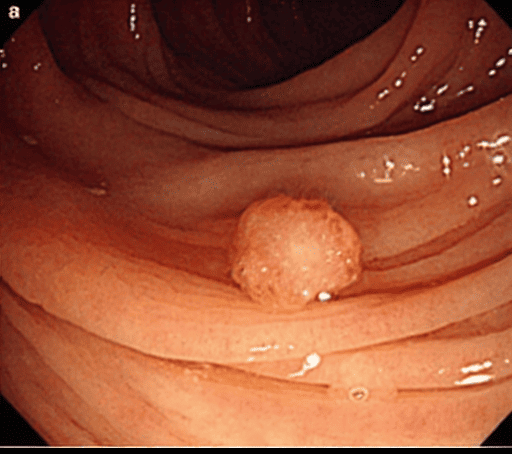} \\
    \multicolumn{2}{c}{\small Remove intestinal polyps.} & \small 0.8/0.8/0.8 & \small 0.2/0.3/0.2 & \small 0.2/0.7/0.5 & \small 0.3/0.7/0.8 & \small 0.1/0.5/0.7 & \small 0.0/0.3/0.3 & \small 0.2/0.6/0.7 \\
    \midrule
    \includegraphics[width=0.1\textwidth]{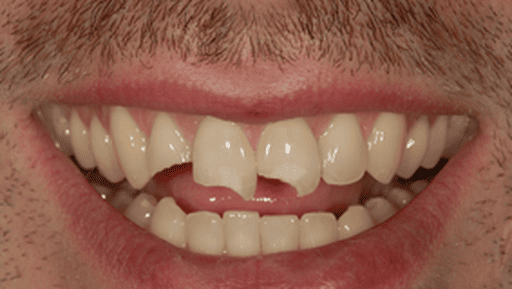} & \includegraphics[width=0.1\textwidth]{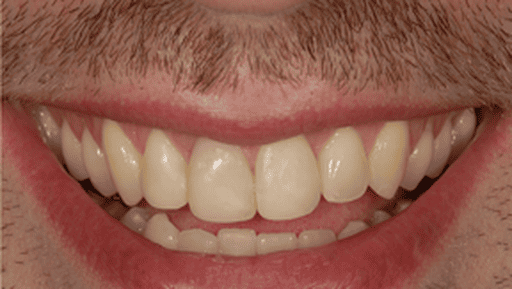} & \includegraphics[width=0.1\textwidth]{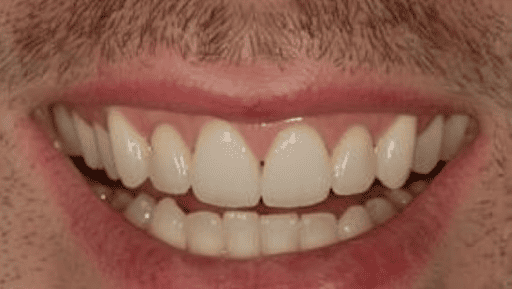} & \includegraphics[width=0.1\textwidth]{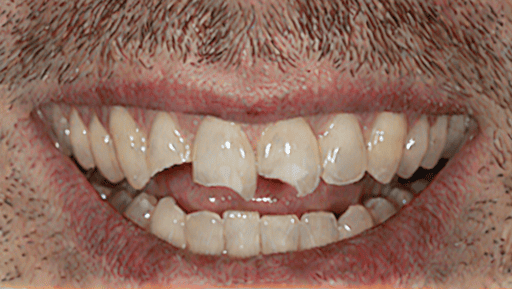} & \includegraphics[width=0.1\textwidth]{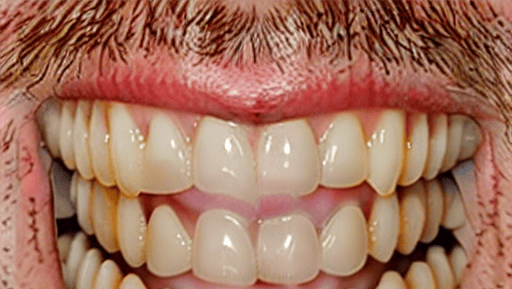} & \includegraphics[width=0.1\textwidth]{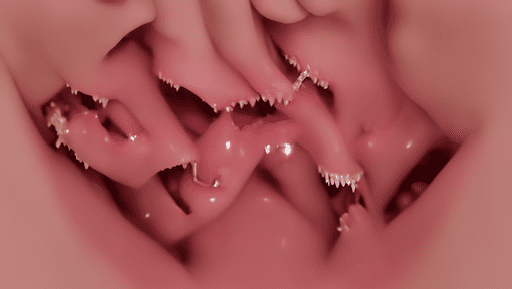} & \includegraphics[width=0.1\textwidth]{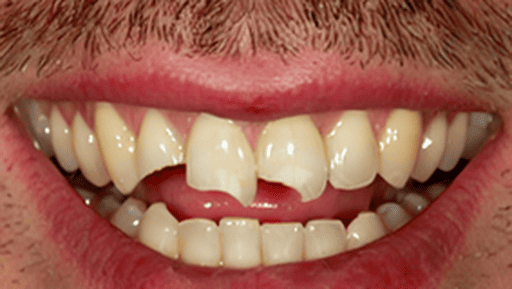} & \includegraphics[width=0.1\textwidth]{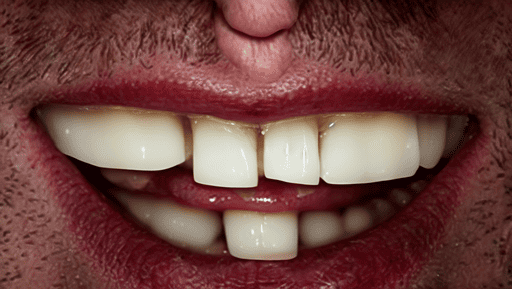} & \includegraphics[width=0.1\textwidth]{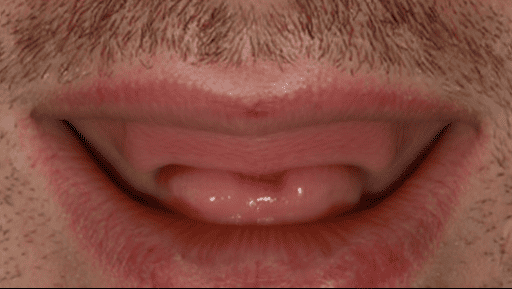} \\
    \multicolumn{2}{c}{\small Fix damaged front teeth.} & \small 0.9/0.8/0.9 & \small 0.0/0.5/0.8 & \small 0.7/0.6/0.5 & \small 0.0/0.0/0.7 & \small 0.2/0.5/0.6 & \small 0.9/0.9/0.7 & \small 0.0/0.4/0.9 \\
    \midrule
    \includegraphics[width=0.1\textwidth]{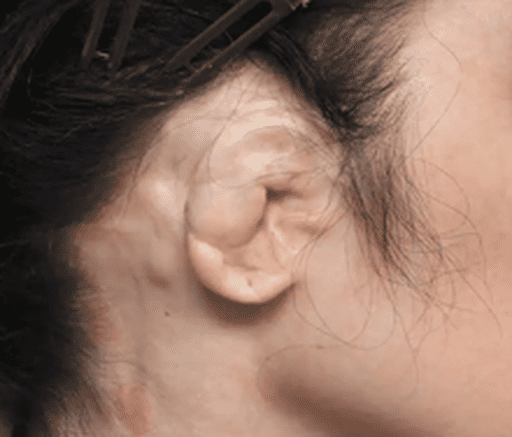} & \includegraphics[width=0.1\textwidth]{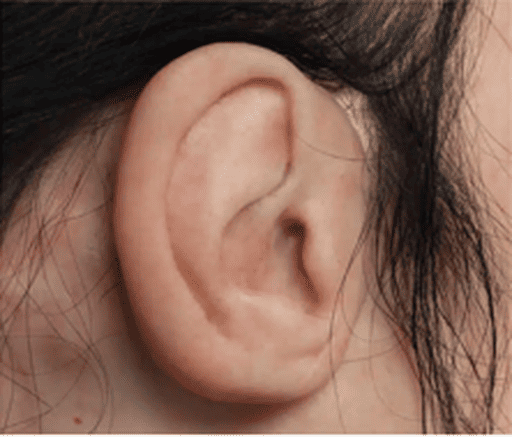} & \includegraphics[width=0.1\textwidth]{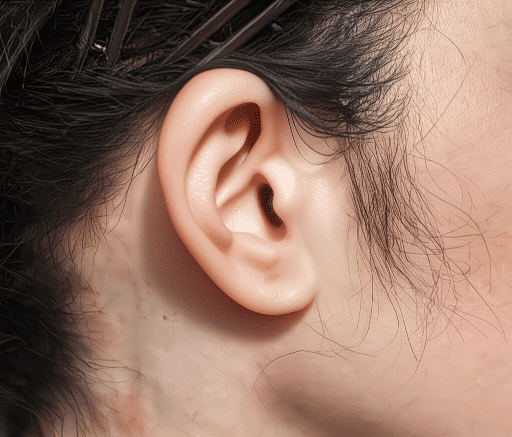} & \includegraphics[width=0.1\textwidth]{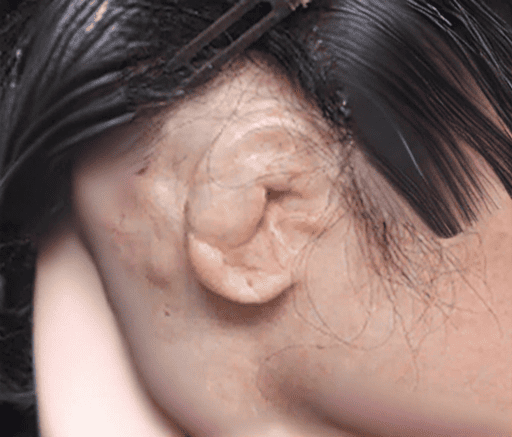} & \includegraphics[width=0.1\textwidth]{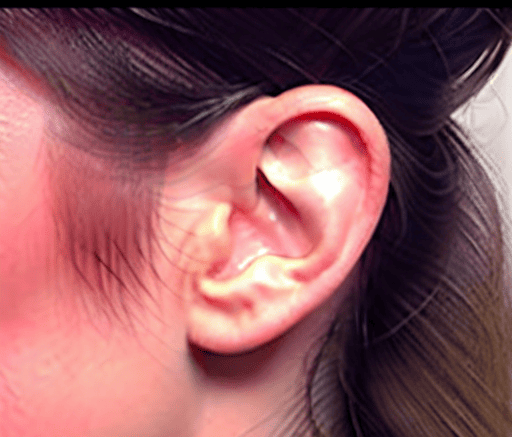} & \includegraphics[width=0.1\textwidth]{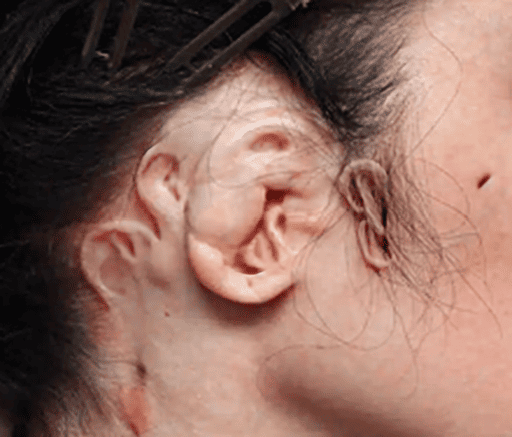} & \includegraphics[width=0.1\textwidth]{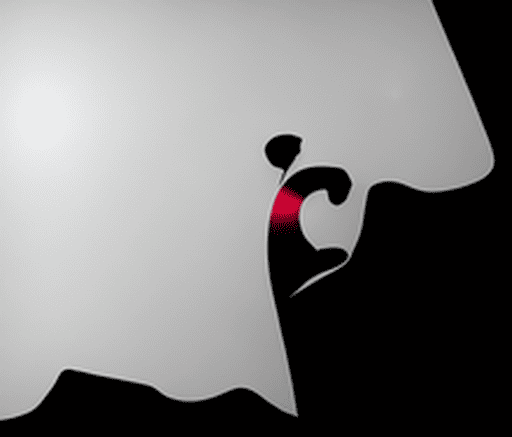} & \includegraphics[width=0.1\textwidth]{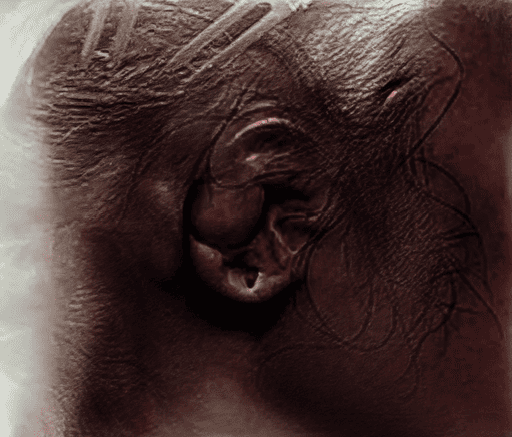} & \includegraphics[width=0.1\textwidth]{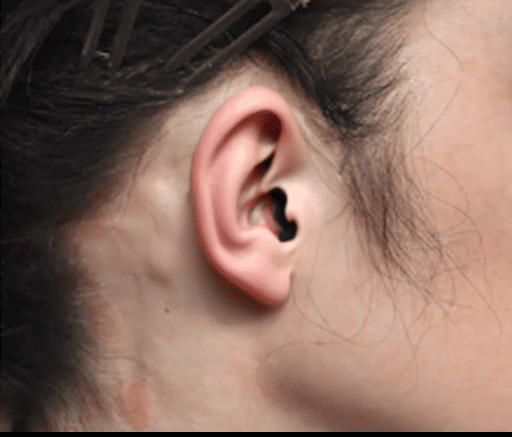} \\
    \multicolumn{2}{c}{\small Reconstruct the ear.} & \small 0.9/0.8/0.6 & \small 0.2/0.4/0.7 & \small 0.7/0.6/0.4 & \small 0.2/0.3/0.9 & \small 0.0/0.3/0.3 & \small 0.0/0.2/0.4 & \small 0.8/0.7/0.9 \\
    \midrule
    \includegraphics[width=0.1\textwidth]{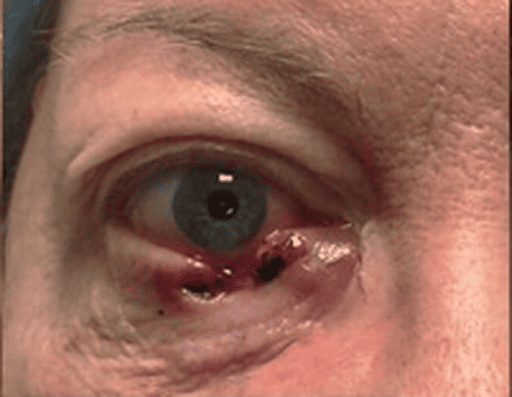} & \includegraphics[width=0.1\textwidth]{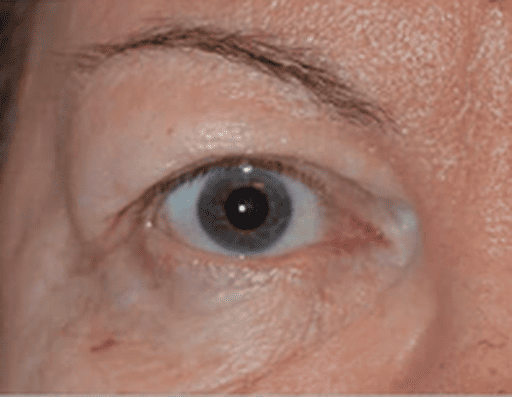} & \includegraphics[width=0.1\textwidth]{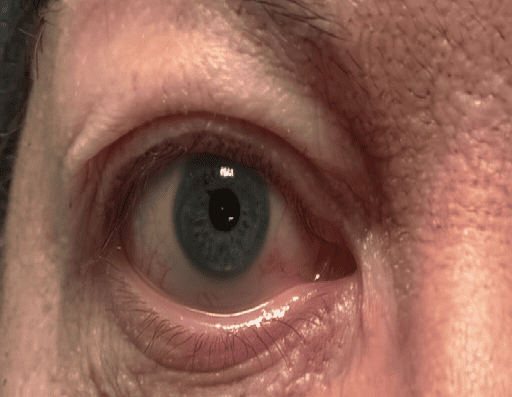} & \includegraphics[width=0.1\textwidth]{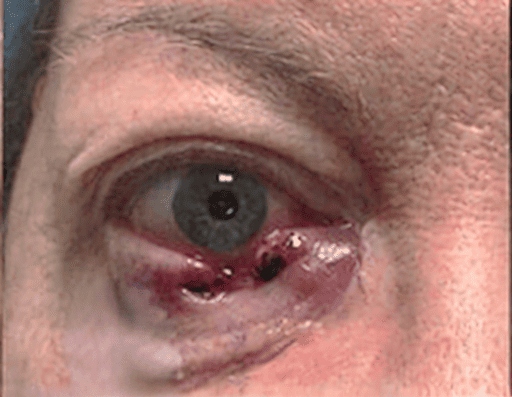} & \includegraphics[width=0.1\textwidth]{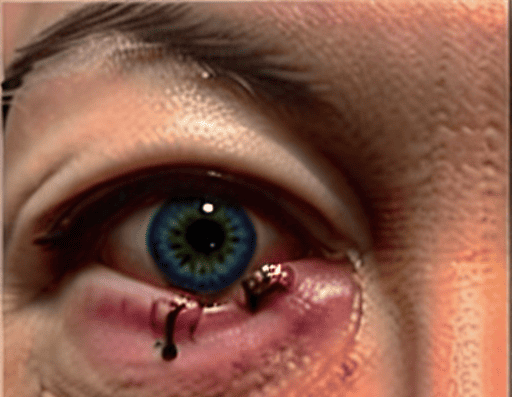} & \includegraphics[width=0.1\textwidth]{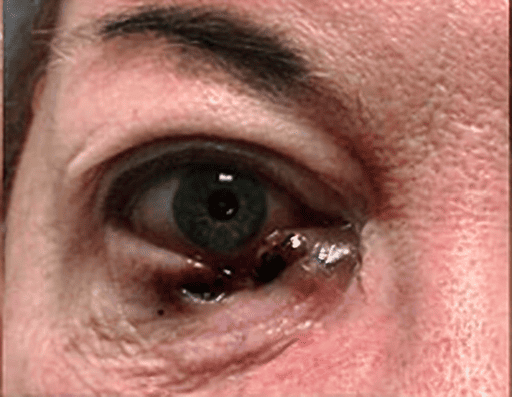} & \includegraphics[width=0.1\textwidth]{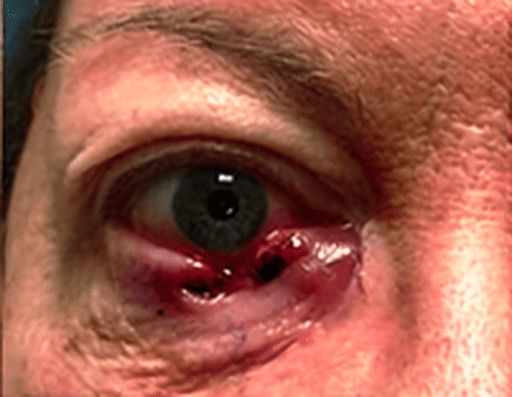} & \includegraphics[width=0.1\textwidth]{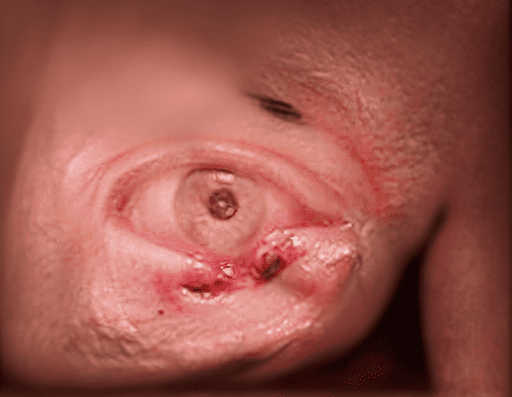} & \includegraphics[width=0.1\textwidth]{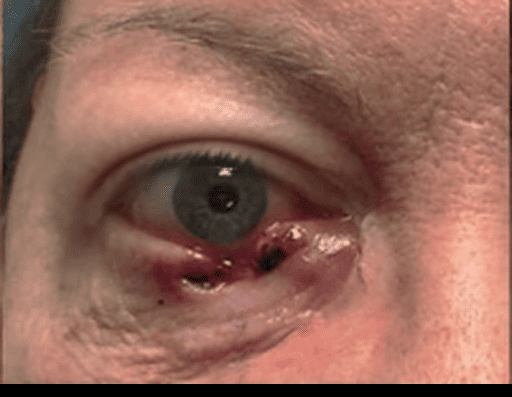} \\
    \multicolumn{2}{c}{\small Reconstruct lower eyelid.} & \small 0.9/0.9/0.6 & \small 0.4/0.5/0.9 & \small 0.7/0.9/0.6 & \small 0.3/0.8/0.9 & \small 0.2/0.9/0.9 & \small 0.4/0.8/0.6 & \small 0.2/0.9/0.7 \\
    \midrule
    \includegraphics[width=0.1\textwidth]{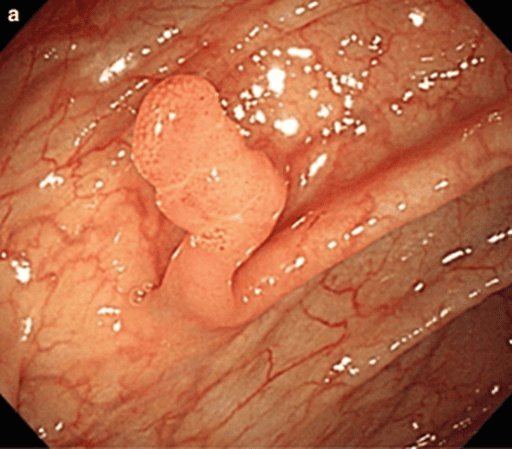} & \includegraphics[width=0.1\textwidth]{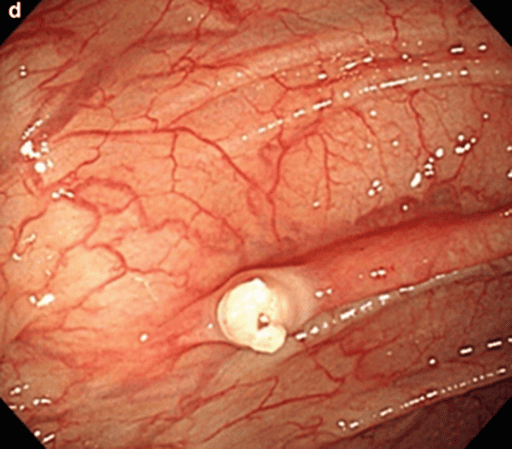} & \includegraphics[width=0.1\textwidth]{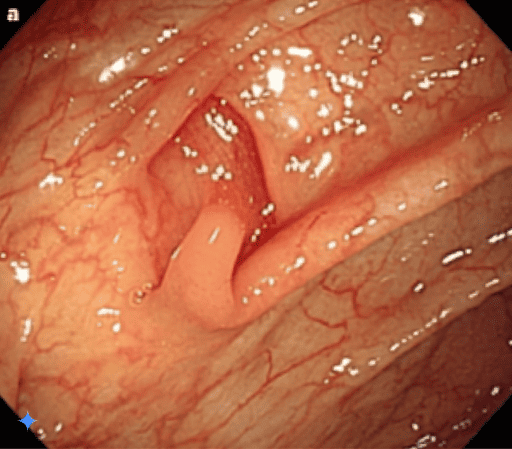} & \includegraphics[width=0.1\textwidth]{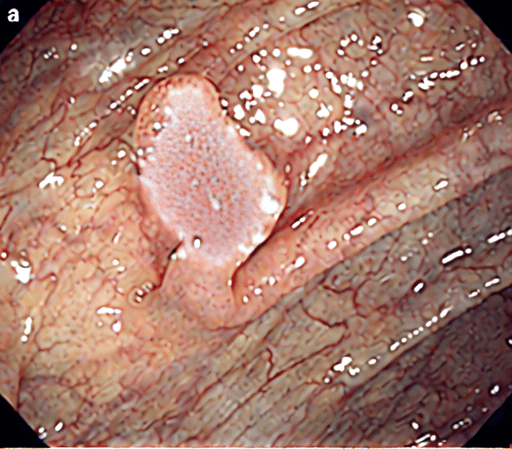} & \includegraphics[width=0.1\textwidth]{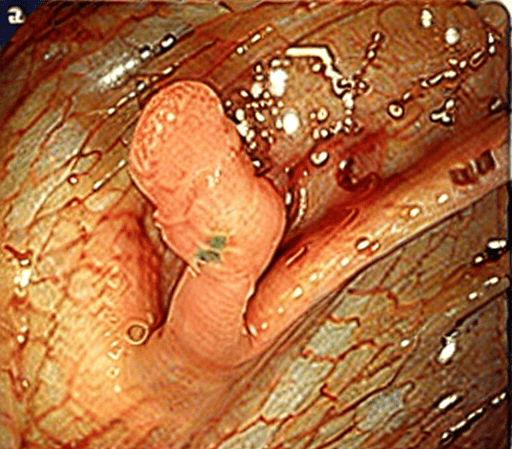} & \includegraphics[width=0.1\textwidth]{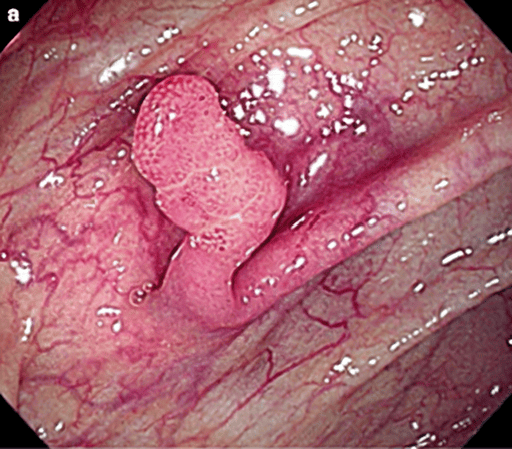} & \includegraphics[width=0.1\textwidth]{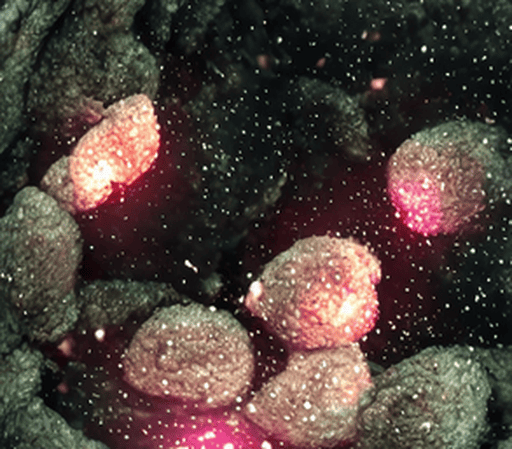} & \includegraphics[width=0.1\textwidth]{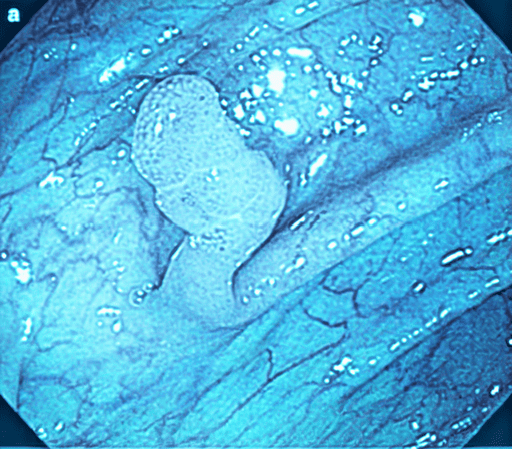} & \includegraphics[width=0.1\textwidth]{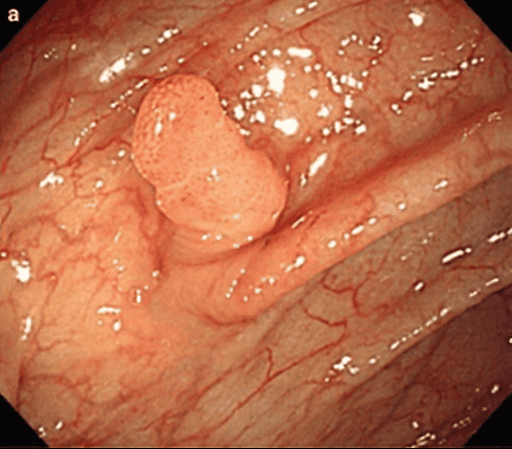} \\
    \multicolumn{2}{c}{\small Remove intestinal adenoma.} & \small 0.8/0.6/0.9 & \small 0.7/0.7/0.7 & \small 0.2/0.5/0.3 & \small 0.0/0.6/0.8 & \small 0.0/0.3/0.2 & \small 0.0/0.4/0.5 & \small 0.0/1.0/0.9 \\
    \midrule
    \includegraphics[width=0.1\textwidth]{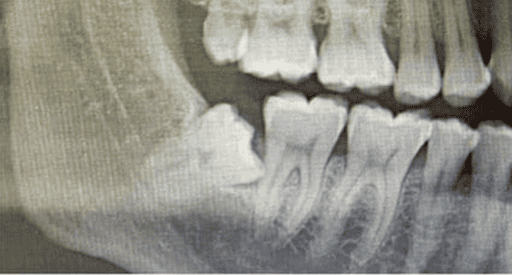} & \includegraphics[width=0.1\textwidth]{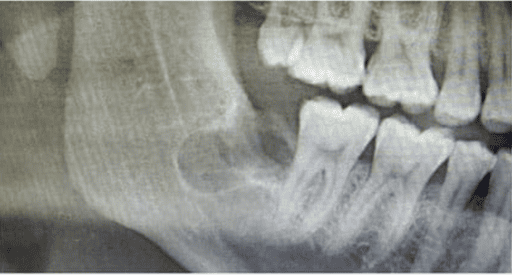} & \includegraphics[width=0.1\textwidth]{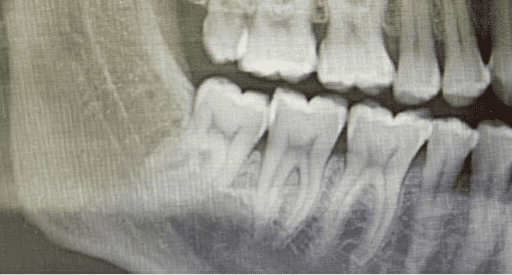} & \includegraphics[width=0.1\textwidth]{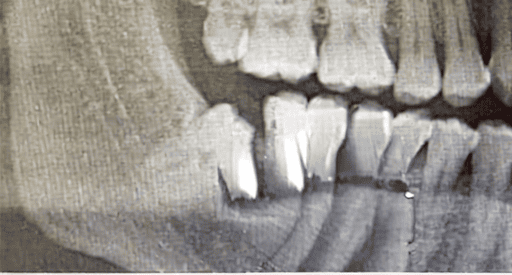} & \includegraphics[width=0.1\textwidth]{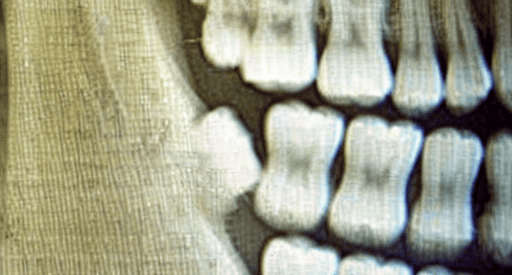} & \includegraphics[width=0.1\textwidth]{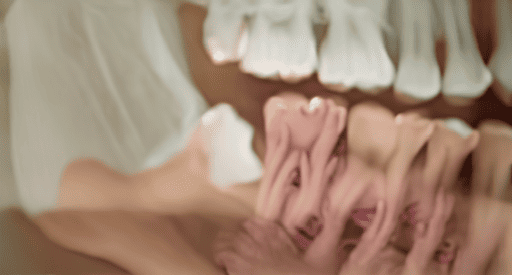} & \includegraphics[width=0.1\textwidth]{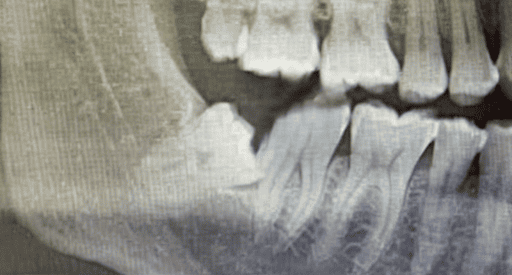} & \includegraphics[width=0.1\textwidth]{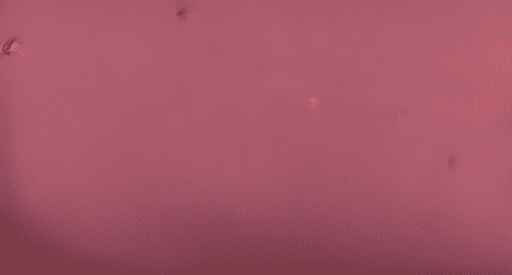} & \includegraphics[width=0.1\textwidth]{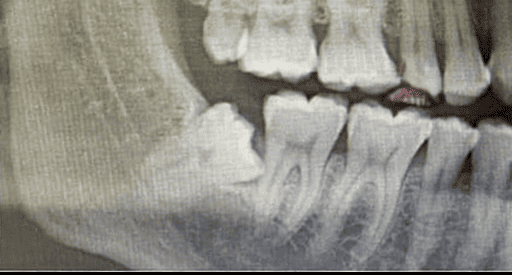} \\
    \multicolumn{2}{c}{\small Remove wisdom teeth.} & \small 0.6/0.7/0.8 & \small 0.5/0.4/0.8 & \small 0.7/0.6/0.4 & \small 0.2/0.1/0.8 & \small 0.7/0.6/0.5 & \small 0.0/0.2/0.5 & \small 0.2/0.6/0.8 \\
    \midrule
    \includegraphics[width=0.1\textwidth]{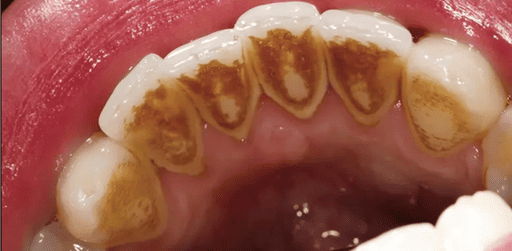} & \includegraphics[width=0.1\textwidth]{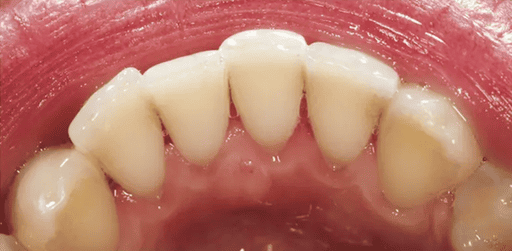} & \includegraphics[width=0.1\textwidth]{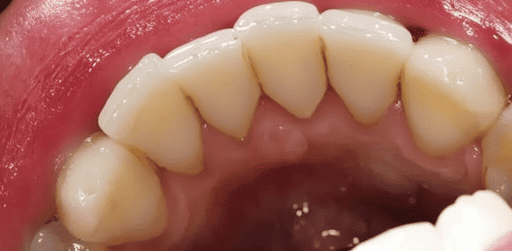} & \includegraphics[width=0.1\textwidth]{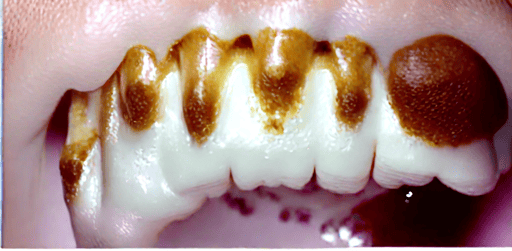} & \includegraphics[width=0.1\textwidth]{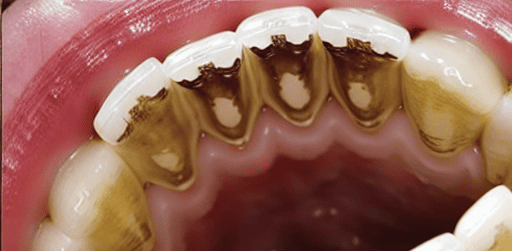} & \includegraphics[width=0.1\textwidth]{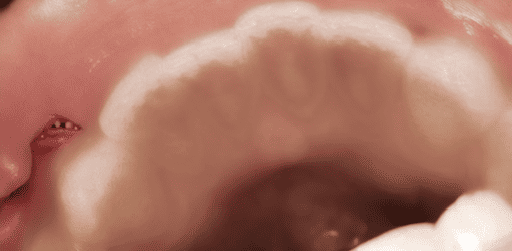} & \includegraphics[width=0.1\textwidth]{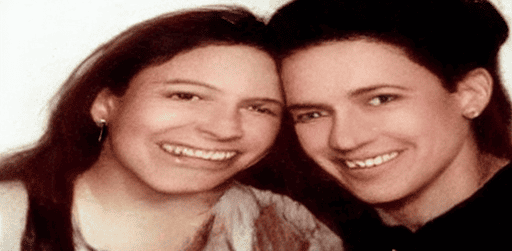} & \includegraphics[width=0.1\textwidth]{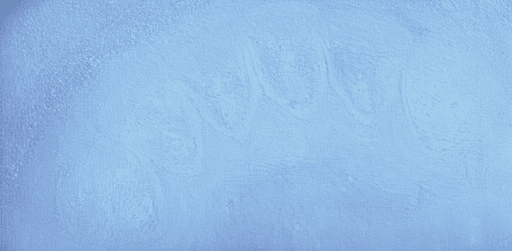} & \includegraphics[width=0.1\textwidth]{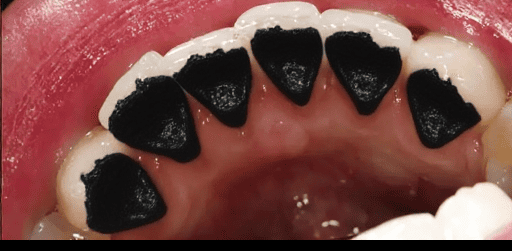} \\
    \multicolumn{2}{c}{\small Remove dental black stains.} & \small 0.9/0.8/0.9 & \small 0.0/0.3/0.8 & \small 0.2/0.5/0.8 & \small 0.8/0.7/0.9 & \small 0.0/0.0/0.7 & \small 0.0/0.2/0.8 & \small 0.1/0.3/0.9 \\
    \midrule
    \end{tabular}
    \caption{Visual comparison of editing results. Each row includes the previous and ground truth images, followed by outputs from seven models. Scores below each output denote EA (Editing Accuracy), VQ (visual quality), and CP (masked SSIM) (all in [0, 1] range).}
    \label{tab:editing_table}
\end{table*}

\clearpage
\section{Learning Paradigm Comparison}
\label{sec:appendix_learning_paradigm}

\vspace{-0.5em}
This section provides a detailed analysis of the two learning paradigms—\textit{fine-tuning} and \textit{in-context learning}—as introduced in Section~\ref{subsec:Learning_Paradigm_Comparison_Results}. Table~\ref{tab:Learning_Paradigms_result} reports results across six representative medical image editing tasks, with Editing Accuracy (EA), Contextual Preservation (CP), and Visual Quality (VQ) reported per setting. Fine-tuning varies training samples per task (0, 4, 8, 16), while in-context learning varies the number of demonstrations (0, 1, 2, 3).

\vspace{-0.5em}
\paragraph{Fine-tuning (InstructPix2Pix).}
Fine-tuning shows consistent gains on structured reconstruction tasks such as \textit{Reconstruct Ear} and \textit{Reconstruct Nose}, where EA steadily improves with more training samples. In contrast, performance on removal tasks (\textit{Remove Wisdom Teeth}, \textit{Remove Moles}) often plateaus or declines, suggesting that excessive supervision can lead to overfitting or degraded semantic precision in fine-grained edits.

\vspace{-0.5em}
\paragraph{In-context Learning (Gemini 2).}
Gemini 2 performs well in the zero-shot setting on several tasks, but shows limited benefit from additional demonstrations. In some cases, performance declines with more shots—for example, EA for \textit{Reconstruct Ear} drops from 0.75 to 0.72 from 0-shot to 3-shot. This suggests difficulty in distinguishing test inputs from exemplars, limiting the model’s ability to generalize in pixel-level editing tasks.

Overall, fine-tuning is more effective when modest amounts of task-specific supervision are available, particularly for structured and spatially consistent reconstruction tasks. In contrast, in-context learning with large-scale models like Gemini offers reasonable generalization in zero-shot settings but does not scale effectively with additional examples. These findings highlight the need for improved prompting strategies or architectural enhancements to enable reliable few-shot medical image editing via in-context learning.

\begin{table*}[h!]
\centering
\renewcommand{\arraystretch}{1.2}
\setlength{\tabcolsep}{8pt}
\textbf{(a) InstructPix2Pix (Finetuning Samples)} \\
\vspace{2mm}
\begin{adjustbox}{max width=\textwidth}
\begin{tabular}{lcccc}
\toprule
\textbf{Task} & \textbf{0s} & \textbf{4s} & \textbf{8s} & \textbf{16s} \\
\midrule
Reconstruct ear & 0.34 / 0.90 / 0.67 & 0.35 / 0.90 / 0.69 & 0.38 / 0.87 / 0.67 & 0.43 / 0.86 / 0.68 \\
Remove wisdom teeth & 0.31 / 0.87 / 0.42 & 0.31 / 0.85 / 0.40 & 0.29 / 0.81 / 0.39 & 0.30 / 0.83 / 0.40 \\
Remove moles & 0.46 / 0.83 / 0.72 & 0.45 / 0.85 / 0.73 & 0.41 / 0.84 / 0.73 & 0.37 / 0.82 / 0.68 \\
Reconstruct nose & 0.68 / 0.88 / 0.77 & 0.70 / 0.86 / 0.74 & 0.75 / 0.86 / 0.75 & 0.77 / 0.87 / 0.75 \\
Remove varicose veins & 0.62 / 0.75 / 0.32 & 0.64 / 0.73 / 0.31 & 0.67 / 0.74 / 0.35 & 0.67 / 0.77 / 0.34 \\
Reconstruct toenails & 0.17 / 0.90 / 0.48 & 0.19 / 0.90 / 0.48 & 0.20 / 0.90 / 0.48 & 0.25 / 0.90 / 0.48 \\
\bottomrule
\end{tabular}
\end{adjustbox}

\vspace{1mm}
\textbf{(b) Gemini 2 (In-Context Samples)} \\
\vspace{1mm}
\begin{adjustbox}{max width=\textwidth}
\begin{tabular}{lcccc}
\toprule
\textbf{Task} & \textbf{0s} & \textbf{1s} & \textbf{2s} & \textbf{3s} \\
\midrule
Reconstruct ear & 0.75 / 0.70 / 0.84 & 0.73 / 0.65 / 0.83 & 0.71 / 0.63 / 0.82 & 0.72 / 0.57 / 0.81 \\
Remove wisdom teeth & 0.70 / 0.81 / 0.79 & 0.72 / 0.78 / 0.80 & 0.70 / 0.77 / 0.79 & 0.70 / 0.73 / 0.77 \\
Remove moles & 0.87 / 0.84 / 0.77 & 0.84 / 0.77 / 0.78 & 0.83 / 0.78 / 0.80 & 0.84 / 0.72 / 0.77 \\
Reconstruct nose & 0.65 / 0.85 / 0.76 & 0.67 / 0.79 / 0.75 & 0.65 / 0.78 / 0.76 & 0.65 / 0.80 / 0.75 \\
Remove varicose veins & 0.73 / 0.77 / 0.63 & 0.73 / 0.76 / 0.65 & 0.74 / 0.75 / 0.66 & 0.73 / 0.75 / 0.67 \\
Reconstruct toenails & 0.81 / 0.83 / 0.80 & 0.82 / 0.80 / 0.82 & 0.84 / 0.82 / 0.81 & 0.82 / 0.80 / 0.80 \\
\bottomrule
\end{tabular}
\end{adjustbox}
\caption{Evaluation results of (a) Finetuned InstructPix2Pix with 0, 4, 8, and 16 finetuning samples per task, and (b) Gemini 2 in-context learning with 0, 1, 2, and 3 in-context samples. Each cell reports Editing Accuracy (EA), Contextual Preservation (CP), and Visual Quality (VQ) in the format: EA / CP / VQ.}
\label{tab:Learning_Paradigms_result}
\end{table*}

\section{Attention Grounding}
\begin{figure*}[htbp]
\centering
\includegraphics[width=\textwidth]{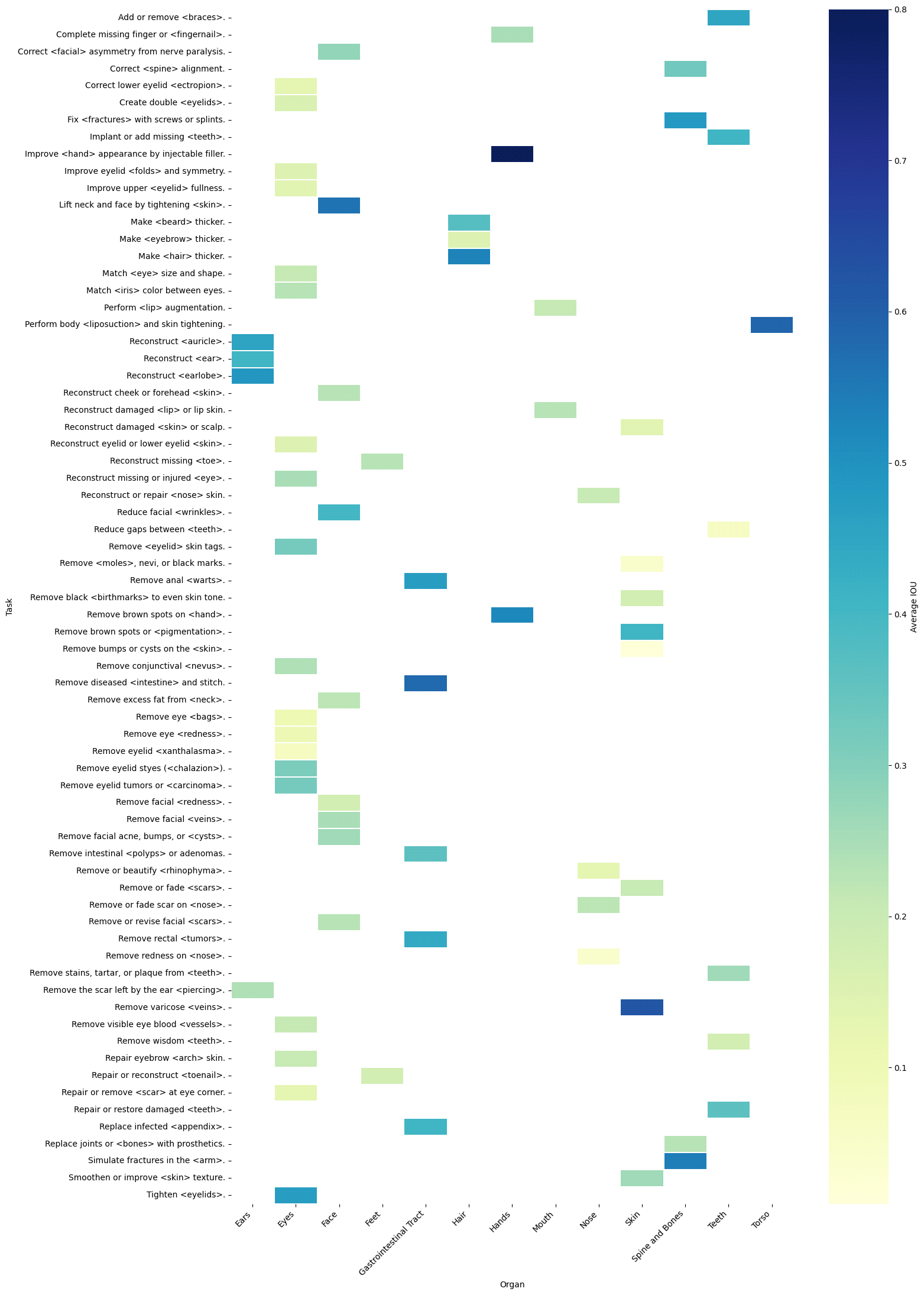}
\caption{Average IOU scores for different tasks across anatomical regions. Concepts are annotated in angle brackets.}
\label{fig:iou_task}
\end{figure*}

Figure~\ref{fig:iou_task} illustrates the detailed average Intersection over Union (IOU) scores across tasks and anatomical regions.

\clearpage
\section{Prompt Templates}
\subsection{GPT-4o Prompt for Generating Change Descriptions}
\label{appendix:gpt4o_prompt_generation}

\textbf{Input:} A concatenated image (previous on the left, changed on the right) and an editing instruction.

\begin{tcolorbox}[
  title=GPT-4o Change Description Prompt,
  colback=gray!3,
  colframe=gray!50,
  breakable,
  enhanced jigsaw,
  sharp corners=south,
  fonttitle=\bfseries,
]

You are a medical expert. Your task is to use your extensive medical knowledge to describe the visual change from the left image to the right image, guided by the clues provided in the editing instruction. Based on these inputs, generate a detailed, precise, and clinically relevant description of the visual change that was applied.

Your response must clearly state:
\begin{itemize}
  \item The image modality
  \item The action performed and its target entity
  \item The expected appearance or anatomical result after the change
\end{itemize}

Format your output as a single formal instruction sentence describing how the right image was derived from the left.

\vspace{1em}
\textbf{Instruction:} \texttt{"<editing\_instruction>"} \\[0.5em]

\textbf{Concatenated image:} \texttt{<base64-encoded image with previous on the left and changed on the right>}
\end{tcolorbox}

\subsection{GPT-4o Judging Prompt (with editing prompt)}
\label{appendix:gpt4o_prompt_prompt}

\textbf{Input:} Previous image, ground truth image, edited image, and editing prompt.

\begin{tcolorbox}[
  title=GPT-4o Judging Prompt,
  colback=gray!3,
  colframe=gray!50,
  breakable,
  enhanced jigsaw,
  sharp corners=south,
  fonttitle=\bfseries,
]

I have an image editing task. Here's the editing prompt: \\
\texttt{"<editing\_prompt>"} \\[0.5em]

Here is the input image: \\
\texttt{<base64-encoded previous image>} \\[0.5em]

Here is the ground truth image: \\
\texttt{<base64-encoded ground truth image>} \\[0.5em]

Here is the edited image: \\
\texttt{<base64-encoded edited image>} \\[0.5em]

You are good at evaluating image editing. Evaluate the edit using the following steps:

\vspace{0.5em}
\textbf{Step 1: Visual Difference Description} \\
Compare the input and edited images. Describe all visible differences between them, including:
\begin{itemize}
  \item Additions, removals, or modifications of visual elements.
  \item Emphasize the extent of the changes and specify which anatomical regions were affected.
\end{itemize}
List the differences clearly, item by item.

\vspace{0.5em}
\textbf{Step 2: Evaluation (Three Scores)} \\
Provide the following scores with brief justifications:

\begin{itemize}
  \item \textbf{1. Editing Accuracy (0--10):} \\
  Score strictly based on alignment with the editing prompt. You may use the ground truth image as a reference, though it may not be pixel-aligned. Deduct points for inaccuracies or missing elements.

  \item \textbf{2. Visual Quality (0--10):} \\
  Compare the overall visual quality of the edited image to the original. Consider clarity, sharpness, blur, artifacts, realism, and consistency.
\end{itemize}

\vspace{0.5em}
\textbf{Please respond in the following format:}
\begin{itemize}
  \item \textbf{Visual Difference Description:}
  \begin{itemize}
    \item 1. ...
    \item 2. ...
    \item ...
  \end{itemize}
  \item \textbf{Editing Accuracy:} [Score]/10, [Reasons]
  \item \textbf{Visual Quality:} [Score]/10, [Reasons]
\end{itemize}
\end{tcolorbox}

\subsection{GPT-4o Judging Prompt (with change description)}
\label{appendix:gpt4o_prompt_description}

\textbf{Input:} Previous image, ground truth image, edited image, and change description.

\begin{tcolorbox}[
  title=GPT-4o Judging Prompt,
  colback=gray!3,
  colframe=gray!50,
  breakable,
  enhanced jigsaw,
  sharp corners=south,
  fonttitle=\bfseries,
]

I have an image editing task. Here's the description of the expected change: \\
\texttt{"<change\_description>"} \\[0.5em]

Here is the input image: \\
\texttt{<base64-encoded previous image>} \\[0.5em]

Here is the ground truth image: \\
\texttt{<base64-encoded ground truth image>} \\[0.5em]

Here is the edited image: \\
\texttt{<base64-encoded edited image>} \\[0.5em]

You are good at evaluating image editing. Evaluate the edit using the following steps:

\vspace{0.5em}
\textbf{Step 1: Visual Difference Description} \\
Compare the input and edited images. Describe all visible differences between them, including:
\begin{itemize}
  \item Additions, removals, or modifications of visual elements.
  \item Emphasize the extent of the changes and specify which anatomical regions were affected.
\end{itemize}
List the differences clearly, item by item.

\vspace{0.5em}
\textbf{Step 2: Evaluation (Three Scores)} \\
Provide the following scores with brief justifications:

\begin{itemize}
  \item \textbf{1. Editing Accuracy (0--10):} \\
  Score strictly based on alignment with the editing prompt. You may use the ground truth image as a reference, though it may not be pixel-aligned. Deduct points for inaccuracies or missing elements.

  \item \textbf{2. Visual Quality (0--10):} \\
  Compare the overall visual quality of the edited image to the original. Consider clarity, sharpness, blur, artifacts, realism, and consistency.
\end{itemize}

\vspace{0.5em}
\textbf{Please respond in the following format:}
\begin{itemize}
  \item \textbf{Visual Difference Description:}
  \begin{itemize}
    \item 1. ...
    \item 2. ...
    \item ...
  \end{itemize}
  \item \textbf{Editing Accuracy:} [Score]/10, [Reasons]
  \item \textbf{Visual Quality:} [Score]/10, [Reasons]
\end{itemize}
\end{tcolorbox}



\end{document}